**Cyclic Directed Probabilistic Graphical Model: A Proposal Based on Structured Outcomes**

*Oleksii Sirotkin*

Masters in software engineering

Graduate student, G.E. Pukhov Institute for Modeling in Energy Engineering

E-mail: cabemailbox@gmail.com

https://orcid.org/0000-0002-6330-6096

**Abstract:** In the process of building (structural learning) a probabilistic graphical model from a set of observed data, the directional, cyclic dependencies between the random variables of the model are often found. Existing graphical models such as Bayesian and Markov networks can reflect such dependencies. However, this requires complicating those models, such as adding additional variables or dividing the model graph into separate subgraphs. Herein, we describe a probabilistic graphical model—probabilistic relation network— that allows the direct capture of directional cyclic dependencies during structural learning. This model is based on the simple idea that each sample of the observed data can be represented by an arbitrary graph (structured outcome), which reflects the structure of the dependencies of the variables included in the sample. Each of the outcomes contains only a part of the graphical model structure; however, a complete graph of the probabilistic model is obtained by combining different outcomes. Such a graph, unlike Bayesian and Markov networks, can be directed and can have cycles. We explored the full joint distribution and conditional distribution and conditional independence properties of variables in the proposed model. We defined the algorithms for constructing of the model from the dataset and for calculating the conditional and full joint distributions. We also performed a numerical comparison with Bayesian and Markov networks. This model does not violate the probability axioms, and it supports learning from observed data. Notably, it supports probabilistic inference, making it a prospective tool in data analysis and in expert and design-making applications.

## 1. Introduction and motivation

Probabilistic graphical models are extensively used in the field of data analysis, machine learning, and artificial intelligence [1]. In fact, such models are a set of random variables (graph nodes), whose values reflect (modeling) some important properties in the world, for example in case we model a coin toss experiment our variables will reflect a coins and values will reflect the side of the coin i.e., head or tail. Variables are connected by a probabilistic relationship (graph edges), also known as joint probability distributions (PD), and edges reflect the relationships of the properties of the world [2]. The edges of the graph can be undirected and directed. Additionally, the directionality of the edges allows the model to capture the type of world property relationships, such as causality.

Presently, two types of probabilistic graph models—Bayesian and Markov networks— as well as their variations, are widely used in practice [1, 2, 3]. Bayesian network is a directed acyclic graph (DAG) and can model directed relationships but cannot directly model cyclic relationships. A Markov network (also known as Markov Random Fields) is a undirected cyclic graph (UCG); hence, it can model cyclic relationships, but only if they are in canonical form, while it cannot model directed relationships [4].

However, as practice shows, in many cases, relationships between the properties of the world can be directed and cyclic [5]; hence, a directed cyclic graph (DCG) is required to capture them. This is especially evident in the process of building a network structure from a dataset (structural learning). In general, Markov and Bayesian networks can model such relationships; however, this will make the model complex. For example, in Markov networks, additional variables encoding the type of relationship can be used for this purpose. Alternatively, in Bayesian networks, this can be achieved by representing DCG in the form of several DAGs [6].

Other researchers have also noticed the difficulties associated with modeling cyclic directional relationships and have offered their own solutions. For example, in a previous study [7], Kłopotek M. proposed to interpret the Bayesian network in terms of the Markov process, thereby allowing the existence of cyclic dependencies in Bayesian network graph. The problem with this approach is that the Bayesian network is represented as a Markov chain, which can potentially be infinite. In another study [5], authors proposed to extend existing learning algorithms and causal inference algorithms in DAG models (such as Bayesian networks) by adding necessary support for the cyclic dependencies between random variables. In contrast, our work focuses more on developing a new model rather than making improvement to algorithms of known models. Certain papers [8, 9] show the application of Markov networks for modeling directed cyclic dependencies. However, since a Markov network is, by definition, a UCG, this requires additional complication of its graphical structure.

The main objective of this work is to define a graphical probabilistic model capable of directly capturing the complex relationships between the properties of the world; predominantly, directions and cyclic relationships. However, in the course of the work, we found a way to classify relationships into a general type that includes (but is not limited to) directionality. We have named this model the probabilistic relation network (PRN) to emphasize the central meaning of relationships contained in it.

Such a model, on one hand, can be built from large datasets without excessive complexity. On the other hand, it will be capable of reflecting the relations of the properties of the world more accurately, which will increase the accuracy and adequacy of the probabilistic inferences in practical applications.

The rest of our work is structured as follows: In section 2, we will first describe the idea behind the PRN, then, we will give its formal definition. Further, we will consider the properties of full joint and conditional PDs. Thereafter, we will additionally describe the special case of a PRN—a factorized network; finally, we will show with a simple example how the model can be used in practice. In section 3, we describe, implement and test the algorithms required for building a PRN, computation of conditional PD, and computation of full-joint PD for a factorized network. In section 4, we will show how Bayesian and Markov networks can be represented using a PRN and perform numeric comparisons on a few simple examples. In section 5, we will summarize our work and share ideas on how it can be continued in the future.

## 2. Definition of probabilistic relxation network

At the core of the PRN lays the following trivial idea: The outcome of some experiments can be represented not just by a point in sample space [10] but also by a graph reflecting the outcome's structure. In the graph, we can represent the outcome as a set of probabilistic events, which will be vertices, and represent the relations between these events as edges. Additionally, each vertex (event) will belong to a random variable, which together forms a relation network.

For example, let us imagine an experiment in which we simultaneously toss a pair of coins, say 25¢ and 50¢ (Figure 1). Suppose we are interested not only in how the coins fall out but also in what order they touch the ground. As such, each coin toss result can be represented by a graph of a pair of vertices, each of which is the value "*head*" or "*tail*," belonging to the random variable "*25¢*" or "*50¢*." Additionally, the directed edge between the values is the order the coins touched the ground (Figure 2). Notably, this experiment can also be easily modeled using a Markov network, but this would require an additional variable with the values "*25¢ then 50¢*" and "*50¢ then 25¢*."

Next, we rigorously define a PRN and describe some of its properties.

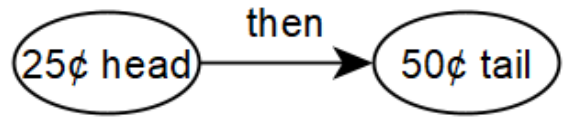

Fig. 2 Example of an outcome of the order of fall of two coins

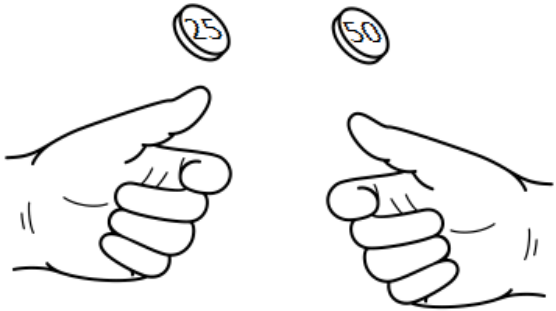

Fig. 1 Two-coin toss experiment

### 2.1. Definition of sample space Ω

The result of some experiment or trial will be called an **outcome** [11]. In our case, each outcome is a connected graph of arbitrary values/events and the relations between them.

**Definition**: *An outcome is a connected graph $\omega = \langle e, r \rangle$. The vertices of which are arbitrary events $e \in \boldsymbol{e}$, belonging to some measurable space $(\boldsymbol{E}, \mathcal{G})$ and $e(\omega) \in \mathcal{G}$, where $e(\omega)$ is the set of $e$ included in $\omega$, and the edges are typed relations $(e_1 \cap^r e_2) \ni \boldsymbol{r}$ of some arbitrary $r$ from the set of possible types of relations $\boldsymbol{R}$.*

Notably, this definition does not impose any restrictions on the type of values $e$ (vertices) and relations $r$ (edges).

Each outcome can have a certain probability ranging from 0 to 1. In this study, we use the frequency interpretation of probability (frequentist probability approach) [12], but other interpretations are also possible.

**Definition**: *The probability of an outcome is defined as*

$$P(\omega) = \lim_{|T| \to \infty} \frac{|\omega(T)|}{|T|}, (1)$$

*where $|T|$ denotes the total number of experiments performed and $|\omega(T)|$ denotes the number of experiments with a specific outcome $\omega$. Moreover, $\omega(T) \subseteq T$.*

We can now define a **probability space** [10] as the set of all possible outcomes (i.e., all possible connected graphs).

**Definition**: *A probability space is a triple $(\boldsymbol{\Omega}, \mathcal{F}, P)$, where $\boldsymbol{\Omega}$ denotes the set of connected graphs $\omega$ that can be formed on the vertex set $\boldsymbol{E}$ using the edges set $\boldsymbol{R}$, the event set $\mathcal{F}$ is the same as the vertex set $\boldsymbol{E}$, and the event probability function $P: \boldsymbol{E} \to \mathbb{R}[0,1]$ is*

$$P(e) = \sum_{\{\omega \in \Omega | e(\omega) \ni e\}} P(\omega), (2)$$

*where $\{\omega \in \boldsymbol{\Omega} | e(\omega) \ni e\}$ is the set of outcomes $\omega$ that include an event $e$.*

We can count the possible number of outcomes in $\boldsymbol{\Omega}$ using the formula

$$|\boldsymbol{\Omega}| = \left((|\boldsymbol{R}| + 1)^{\frac{|\boldsymbol{E}| * (|\boldsymbol{E}| - 1)}{2}}\right) + |\boldsymbol{E}|, (3)$$

where $|\boldsymbol{R}|$ denotes the number of relations and $|\boldsymbol{E}|$ denotes the number of events.

### 2.2. Definition of random variable $V: \boldsymbol{E}$

As in other probabilistic graphical models [2], the vertices of a PRN are **random variables**.

Probably, as in other models, a PRN includes random variables of any type (discrete, continuous, and mixed). However, this statement requires further research. In this study, we considered only discrete random variables.

**Definition**: *A random variable $V$ is a measurable function $V: \Omega \to E$, where $\Omega$ denotes a probability space and $E$ denotes some measurable space, such that*

$$E \neq \emptyset \quad (4)$$

$$\forall e_1 \in E, e_2 \in E, \omega \in \Omega \; \nexists \; e(\omega) \ni e_1 \cap e_2; \; e_1 \neq e_2 \quad (5)$$

$$\forall e_1 \in E_1, e_2 \in E_2 \; \nexists \; e_1 = e_2. \quad (6)$$

In other words, a random variable $V$ has a domain $E$ comprising the set of vertices $e \in E$. Moreover, the probability space $\Omega$ should not contain graph $\omega$ that includes vertices $e$ that belong to the domain $E$ of the same $V$. Additionally, the vertex $e$ can belong to one and only one domain $E$ (i.e., it can belong to only one variable $V$).

Notably, in the probability space $(\Omega, \mathcal{F}, P)$, we can define arbitrarily many random variables $V_1, V_2, \ldots, V_n$, and each of them can have its arbitrary domain $E_1, E_2, \ldots, E_n$.

In the model under study, we allow the existence of outcome $\omega$ that does not include any value of some variable $V$. In other words, we allow an experiment whose result $\omega$ does not provide any relevant information about the variable $V$. To reflect this case in the model and make the model consistent, we assume that each variable contains an **unobservable value**.

**Definition**: *An unobserved value $u$ is a value that is necessarily included in the domain $E$ of the variable $V$ but not included as a vertex in any outcome $\omega \in \Omega$.*

Notably, the value $u$ can be any value from $E$. It can be either chosen from those already defined or added specially.

The PD of the values of a variable $V$ is the set of probabilities of the values/events $e$ included in its domain $E$, except for the value $u$, the probability of which can be interpreted as "the probability of nonobservation of the variable $V$."

**Definition**: *We extend the probability function $P: E \to \mathbb{R}[0,1]$ for a random variable $V$ as*

$$P(e) = \begin{cases} \sum_{\{\omega \in \Omega \mid e(\omega) \ni e\}} P(\omega); & e \neq u \\ 1 - \sum_{\{e \in E \mid e \neq u\}} P(e); & e = u \end{cases}, \quad (7)$$

where $\{\omega \in \mathbf{\Omega} | e(\omega) \ni e\}$ is the set of outcomes $\omega$ that includes event $e$ and $\{e \in \mathbf{E} | e \neq u\}$ is the set of values of variable $V$ excluding the value $u$.

Accordingly, the PD of a variable is

$$P(V) = \{P(e) \mid e \in \mathbf{E}\}. \quad (8)$$

Additionally, the sum of the probabilities of all values will always be equal to 1 (see Theorem A):

$$\sum_{e \in \mathbf{E}} P(e) = 1. \quad (9)$$

In practice, we are interested in the PD of $V$ without considering the unobserved value $u$. We can obtain it via a simple normalization:

$$\dot{P}(V) = \left\{ \frac{P(e)}{\sum_{e \in \mathbf{E}} P(e)} \mid e \in \mathbf{E}, e \neq u \right\}. \quad (10)$$

### 2.3. Definition of relation graph $\mathbf{\Theta}$

Now, we have the necessary components to define a PRN.

**<u>Definition</u>**: *A PRN $\mathbf{\Theta}$ is a graph $\langle \mathbf{V}, \mathbf{R}, \mathbf{O} \rangle$, where the set of vertices $\mathbf{V}$ is the set of random variables $V: \mathbf{E}$ on the probability space $(\mathbf{\Omega}, \mathcal{F}, P)$, $\mathbf{R}$ is the set of relation types between values of $V$, and $\mathbf{O} = \{\omega \in \mathbf{\Omega} | P(\omega) \neq 0\}$. Moreover, a pair of vertices $V_1, V_2 \in \mathbf{V}$ are connected by an edge if and only if there exists $\omega \in \mathbf{O}$ containing the relation $e_1 \cap^r e_2$, where $e_1 \in \mathbf{E}_1$ and $e \in \mathbf{E}_2$.*

In essence, we can think of a PRN as simply a set of outcomes $\omega$ with nonzero probability, whose vertex values $e$ are combined into variables $V: \mathbf{E}$. The edges of the network simply show some probabilistic dependence between vertices $V_1$ and $V_2$.

Notably, this definition does not impose any restrictions on the types of edges (including directed or not) or on the graph's structure. It also admits the existence of $\mathbf{\Theta}$ with no outcomes in it (i.e., $\mathbf{O} = \emptyset$); in this case, the PD of each variable $V$ will be concentrated in the unobserved value $u$ [i.e., $P(u) = 1$ and $\forall e \in \mathbf{E}: P(e) = 0$]. Additionally, a pair of vertices $\dot{V}_1: \dot{\mathbf{E}}_1, \dot{V}_2: \dot{\mathbf{E}}_2 \in \mathbf{V}$ of $\mathbf{\Theta}$ are independent (i.e., $\dot{V}_1 \perp\!\!\!\perp \dot{V}_2$) if there is no outcome $\omega$ such that $\dot{\mathbf{E}}_1, \dot{\mathbf{E}}_2 \subseteq e(\omega)$. In other words, $\dot{V}_1 \perp\!\!\!\perp \dot{V}_2$ if there is no path between them. Meanwhile, nonnormalized $V_1$ and $V_2$ are dependent, via the unobserved value $u$.

Assuming that each variable $V$ has the same number of values (i.e., $\forall V_1: E_1, V_2: E_2 \in \mathbf{V} \nexists: |E_1| \neq |E_2|$), we can calculate the maximum number of outcomes as follows:

$$|\mathbf{O}| = \sum_{k=1}^{|\mathbf{V}|} \left( \binom{|\mathbf{V}|}{k} * C_k * (|\mathbf{E}| - 1)^k \right), (11)$$

where

$$C_k = (|\mathbf{R}| + 1)^{\frac{k*(k-1)}{2}} - \frac{\sum_{i=1}^{k-1} \left( i * \binom{k}{i} * (|\mathbf{R}| + 1)^{\frac{(k-i)*(k-i-1)}{2}} * C_i \right)}{k}, (12)$$

$|\mathbf{R}|$ denotes the number of relation types, $|\mathbf{V}|$ denotes the number of vertices in $\mathbf{\Theta}$, and $|\mathbf{E}|$ denotes the number of values at each vertex including the unobserved $u$.

### 2.4. PDs $\widehat{\mathbf{\Theta}}$ and $\mathbf{\Theta}|\breve{\omega}$ in relation graph

In many cases, we will work with the probabilities of specific outcomes, so we introduce the concept of **outcome PD**.

**Definition**: *We define an outcome PD as follows:*

$$P(\mathbf{\Theta}) = \{P(\omega) \mid \omega \in \mathbf{O}\}. (13)$$

A **full joint PD** is the set of probabilities of all combinations of values of all variables, in the context of a PRN $\mathbf{\Theta}$, i.e., a set of outcomes, each of which includes one value from each variable $V$ constituting $\mathbf{\Theta}$.

**Definition**: *A PRN containing a full joint PD is given by*

$$\widehat{\mathbf{\Theta}} = \langle \widehat{\mathbf{V}}, \widehat{\mathbf{R}}, \{\omega \in \widehat{\mathbf{O}} \mid V(\omega) = \widehat{\mathbf{V}}\} \rangle, (14)$$

*where $V(\omega)$ denotes the set of variables whose values are included in the outcome $\omega$.*

In this case, the full joint PD is

$$P(\widehat{\mathbf{V}}) = \{P(\widehat{V}) \mid \widehat{V} \in \widehat{\mathbf{V}}\}. (15)$$

Notably, $\widehat{\mathbf{\Theta}}$ need not contain outcomes for all possible combinations of values but only outcomes with nonzero probability.

A **conditional PD** can be represented as a subset of outcomes from $\mathbf{\Theta}$ that is consistent with some observed outcome.

**Definition**: *A PRN containing a conditional PD is given by*

$$\mathbf{\Theta}|\breve{\omega} = \langle \mathbf{V}, \mathbf{R}, \{\omega \in \mathbf{O} \mid (V(\breve{\omega}) \cap V(\omega) = \emptyset) \bigvee \left( e(\breve{\omega}) \subseteq e(\omega) \bigwedge r(\breve{\omega}) \subseteq r(\omega) \right) \} \rangle, (16)$$

where $r(\omega)$ denotes the set of relations included in the outcome (edges) and $e(\omega)$ denotes the set of values included in the outcome (outcome vertices).

From a computational perspective, when converting $\boldsymbol{O}$ to $\boldsymbol{O}|\breve{\omega}$, PDs should be normalized as follows:

$$P(\boldsymbol{O}|\breve{\omega}) = \left\{\frac{P(\omega)}{Z} \mid \omega \in \boldsymbol{O}|\breve{\omega}\right\}, (17)$$

where $Z$ denotes the normalization constant.

We can convert $\boldsymbol{O}|\breve{\omega}$ to a network with full joint PD $\widehat{\boldsymbol{O}}|\breve{\omega}$ and then obtain conditional PDs of all variables

$$P(\widehat{\boldsymbol{V}}|\breve{\omega}) = \{P(\widehat{V}|\breve{\omega}) \mid \widehat{V} \in \widehat{\boldsymbol{V}}\}. (18)$$

## 2.5. Factorized relation graph $\widetilde{\boldsymbol{O}}$

A special case of $\boldsymbol{O}$ is a factorized PRN $\widetilde{\boldsymbol{O}}$, such that all outcomes in $\widetilde{\boldsymbol{O}}.\widetilde{\boldsymbol{O}}$ are divided into groups (factors) $\Phi \subseteq \widetilde{\boldsymbol{O}}$. In this case, the following conditions are met:

$$K_0 \notin \widetilde{\boldsymbol{O}} \ (19)$$

$$\forall \Phi \in \widetilde{\boldsymbol{O}} : \forall \omega \in \Phi : V(\omega) = V(\Phi). (20)$$

In other words, the network $\widetilde{\boldsymbol{O}}$ does not contain empty and intersecting outcomes

$$\forall \omega_1, \omega_2 \in \widetilde{\boldsymbol{O}} : \vec{r}(\omega_1) \cap \vec{r}(\omega_2) = \emptyset, (21)$$

where $\vec{r}(\omega)$ denotes the relation $e_1 \cap^r e_2$ in $\omega$.

Intuitively, the network $\widetilde{\boldsymbol{O}}$ is a set of groups of nonintersecting (having no common edges) outcomes. Each outcome in each group will have the same set of variables $V(\omega)$ (i.e., each group or factor $\Phi$ will contain a full joint PD of variables in it). Additionally, for all $V$ in such a group, $P(u) = 0$, which means that we can simply ignore the unobserved values $u$. A relation network structured as such is essentially a factor graph [13].

A pair of vertices $V_1, V_2 \in \widehat{\boldsymbol{V}}$ of the factorized network $\widetilde{\boldsymbol{O}}$ are conditionally independent (i.e., $V_1 \perp\!\!\!\perp V_2|Z$) if all possible paths between $V_1$ and $V_2$ are blocked by observed vertices $Z \in \boldsymbol{Z} \subseteq \widehat{\boldsymbol{V}}$, and $\widetilde{\boldsymbol{O}}$ does not contain any $\omega$ connecting vertices $V$ adjacent to $Z$ on a blocked path (see Theorem B).

## 2.6. Two-coin toss experiment example

Consider an experiment with a tossing of two coins (Figure 1). In this case, we have

$$E = \{h_1, t_1, h_2, t_2\} \quad (22)$$

$$V = \{V_1, V_2\}, \quad (23)$$

thus one random variable for each coin. The variable domains will be

$$E_1 = \{h_1, t_1, u_1\} \quad (24)$$

$$E_2 = \{h_2, t_2, u_2\} \quad (25)$$

One possible set of outcomes in this experimental setting will be

$$O = \begin{bmatrix} \omega_1 = K_0 \\ \omega_2 = h_1 \\ \omega_3 = t_1 \\ \omega_4 = h_2 \\ \omega_5 = t_2 \\ \omega_6 = h_1 \cap^r h_2 \\ \omega_7 = h_1 \cap^r t_2 \\ \omega_8 = t_1 \cap^r h_2 \\ \omega_9 = t_1 \cap^r t_2 \end{bmatrix} ; (26)$$

these outcomes can be represented graphically (see Figure 3). In this case, the type of relation $r$ can be, e.g., nondirected, if we are only interested in the correlation between coin tosses,

$$r = \text{"both,"} \quad (27)$$

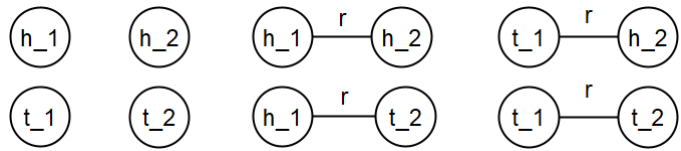

**Fig. 3** O set

or directed, if we want to reflect the sequence relation between coin falls,

$$r_1 = \text{"2 after 1,"} \quad (28)$$

$$r_2 = \text{"1 after 2."} \quad (29)$$

We calculate the total number of possible outcomes $|\Omega|$ when the two relations $R = \{r_1, r_2\}$ are used. In other words, we calculate in how many ways the coins can fall (Formulas 11 and 12)

$$C_1 = (|R| + 1)^{\frac{k*(k-1)}{2}} - \frac{\sum_{i=1}^{k-1}\left(i * \binom{k}{i} * (|R| + 1)^{\frac{(k-i)*(k-i-1)}{2}} * C_i\right)}{k}$$

$$= (|R| + 1)^{\frac{1*(1-1)}{2}} - \frac{\sum_{i=1}^{1-1}\left(i * \binom{1}{i} * (|R| + 1)^{\frac{(1-i)*(1-i-1)}{2}} * C_i\right)}{1}$$

$$= (|R| + 1)^0 - \frac{0}{1} = 1 \, (30)$$

$$C_2 = (|R| + 1)^{\frac{k*(k-1)}{2}} - \frac{\sum_{i=1}^{k-1}\left(i * \binom{k}{i} * (|R| + 1)^{\frac{(k-i)*(k-i-1)}{2}} * C_i\right)}{k}$$

$$= (2 + 1)^1 - \frac{1 * \binom{2}{1} * (2 + 1)^{\frac{(2-1)*(2-1-1)}{2}} * C_1}{2} = 3 - \frac{1 * 2 * 3^0 * 1}{2} = 2 \, (31)$$

$$|O| = \sum_{k=1}^{|V|}\left(\binom{|V|}{k} * C_k * (|E| - 1)^k\right) = \binom{2}{1} * C_1 * (3 - 1)^1 + \binom{2}{2} * C_2 * (3 - 1)^2$$

$$= 2 * 1 * 2 + 1 * 2 * 4 = 4 + 8 = \mathbf{12} \, (32)$$

We toss the coins 10 times. Suppose we obtain a list of results/outcomes

$$T = \begin{bmatrix} K_0 \\ h_1 \\ h_1 \\ t_2 \\ t_1 \\ h_1 \cap^{r_1} h_2 \\ h_1 \cap^{r_1} h_2 \\ h_1 \cap^{r_1} h_2 \\ h_1 \cap^{r_2} h_2 \\ t_1 \cap^{r_1} t_2 \\ h_1 \cap^{r_1} t_2 \end{bmatrix}, (33)$$

the outcomes can be graphically presented as in Figure 4, and the relation network $O$ will look like in Figure 5.

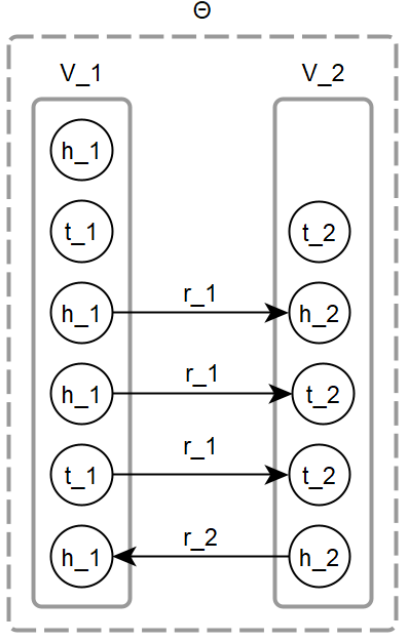

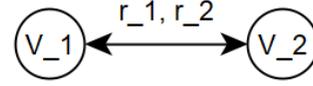

Fig. 5 PRN Θ

Fig. 4 PRN as a set of outcomes

Based on the results of a series of experiments, we can calculate the probabilities of outcomes (Formula 1)

$$P(\Theta) = \begin{cases} P(K_0) = \dfrac{|\{K_0\}|}{|T|} = \dfrac{1}{10} \\ P(h_1) = \dfrac{|\{h_1, h_1\}|}{|T|} = \dfrac{2}{10} \\ P(t_2) = \dfrac{|\{t_2\}|}{|T|} = \dfrac{1}{10} \\ P(t_1) = \dfrac{|\{t_1\}|}{|T|} = \dfrac{1}{10} \\ P(h_1 \cap^{r_1} h_2) = \dfrac{|\{h_1 \cap^{r_1} h_2, h_1 \cap^{r_1} h_2\}|}{|T|} = \dfrac{2}{10} \\ P(h_1 \cap^{r_2} h_2) = \dfrac{|\{h_1 \cap^{r_2} h_2\}|}{|T|} = \dfrac{1}{10} \\ P(t_1 \cap^{r_1} t_2) = \dfrac{|\{t_1 \cap^{r_1} t_2\}|}{|T|} = \dfrac{1}{10} \\ P(h_1 \cap^{r_1} t_2) = \dfrac{|\{h_1 \cap^{r_1} t_2\}|}{|T|} = \dfrac{1}{10} \end{cases}, (34)$$

and the rest of the $12 - 8$ possible outcomes will have zero probability.

Now, we can calculate the probabilities of each value considering them as separate events (Formula 2):

$$P(h_1 \in E) = P(h_1) + P(h_1 \cap^{r_1} h_2) + P(h_1 \cap^{r_2} h_2) + P(h_1 \cap^{r_1} t_2) = \frac{2}{10} + \frac{2}{10} + \frac{1}{10} + \frac{1}{10}$$

$$= \frac{6}{10} \quad (35)$$

$$P(t_1 \in E) = P(t_1) + P(t_1 \cap^{r_1} t_2) = \frac{1}{10} + \frac{1}{10} = \frac{2}{10} \quad (36)$$

$$P(h_2 \in E) = P(h_1 \cap^{r_1} h_2) + P(h_1 \cap^{r_2} h_2) = \frac{2}{10} + \frac{1}{10} = \frac{3}{10} \quad (37)$$

$$P(t_2 \in E) = P(t_2) + P(t_1 \cap^{r_1} t_2) + P(h_1 \cap^{r_1} t_2) = \frac{1}{10} + \frac{1}{10} + \frac{1}{10} = \frac{3}{10}. \quad (38)$$

We can also calculate the conditional probability with respect to the relation type between events, e.g., the probability of $h_1$, if observed that $h_2$ occurred after it ($r_1$ relation)

$$P(h_1|^{r_1}h_2) = \frac{P(h_1 \cap^{r_1} h_2)}{P(h_2)} = \frac{\frac{2}{10}}{\frac{3}{10}} = \frac{2}{3}. \quad (39)$$

We explored the values $E$ as separate events; now, we consider them as values of the variables $V_1: E_1$ and $V_2: E_2$.

In addition to the probabilities of the values calculated earlier, we need to calculate the probabilities of the unobserved values $u_1$ and $u_2$ (Formula 7):

$$P(u_1) = \begin{cases} \sum_{\{\omega \in \Omega | e(\omega) \ni e\}} \cancel{P(\omega); e \neq u} \\ 1 - P(h_1) + P(t_1); \ e = u \end{cases} = \frac{10}{10} - \frac{6}{10} + \frac{2}{10} = \frac{2}{10} \quad (40)$$

$$P(u_2) = \begin{cases} \sum_{\{\omega \in \Omega | e(\omega) \ni e\}} \cancel{P(\omega); e \neq u} \\ 1 - P(h_2) + P(t_2); \ e = u \end{cases} = \frac{10}{10} - \frac{3}{10} + \frac{3}{10} = \frac{4}{10} \quad (41)$$

Thus, we obtain the complete distribution of variables (Formula 8):

$$P(V_1) = \{P(h_1), P(t_1), P(u_1)\} = \left\{\frac{6}{10}, \frac{2}{10}, \frac{2}{10}\right\} \quad (42)$$

$$P(V_2) = \{P(h_2), P(t_2), P(u_2)\} = \left\{\frac{3}{10}, \frac{3}{10}, \frac{4}{10}\right\}. \quad (43)$$

We can also calculate the PD of variables without $u_1$ and $u_2$ values (Formula 10):

$$\dot{P}(V_1) = \left\{\frac{P(h_1)}{P(h_1) + P(t_1)}, \frac{P(t_1)}{P(h_1) + P(t_1)}\right\} = \left\{\frac{\frac{6}{10}}{\frac{6}{10} + \frac{2}{10}}, \frac{\frac{2}{10}}{\frac{6}{10} + \frac{2}{10}}\right\} = \left\{\frac{3}{4}, \frac{1}{4}\right\} \quad (44)$$

$$\dot{P}(V_2) = \left\{ \frac{P(h_2)}{P(h_2) + P(t_2)}, \frac{P(t_2)}{P(h_2) + P(t_2)} \right\} = \left\{ \frac{\frac{3}{10}}{\frac{3}{10} + \frac{3}{10}}, \frac{\frac{3}{10}}{\frac{3}{10} + \frac{3}{10}} \right\} = \left\{ \frac{1}{2}, \frac{1}{2} \right\}. \quad (45)$$

We also check that the probabilities of the values of the variables really sum to 1 (Formula 9)

$$1 = \sum_{e \in E_1} P(e) = \frac{6}{10} + \frac{2}{10} + \frac{2}{10} (45)$$

$$1 = \sum_{e \in E_2} P(e) = \frac{3}{10} + \frac{3}{10} + \frac{4}{10}. \quad (47)$$

The full joint PD $\widehat{\Theta}$ for our example will be (Formula 14)

$$\widehat{\Theta} = \left\langle \begin{array}{c} \widehat{O} = \begin{bmatrix} h_1 \cap {}^{r_1} h_2, h_1 \cap {}^{r_1} t_2, t_1 \cap {}^{r_1} h_2, t_1 \cap {}^{r_1} t_2 \\ h_1 \cap {}^{r_2} h_2, h_1 \cap {}^{r_2} t_2, t_1 \cap {}^{r_2} h_2, t_1 \cap {}^{r_2} t_2 \end{bmatrix} \\ \widehat{R} = \{r_1, r_2\} \\ \widehat{V} = \{V_1: \{u, h_1, t_1\}, V_2: \{u, h_2, t_2\}\} \end{array} \right\rangle. \quad (48)$$

Additionally, if we assume a uniform PD of outcomes, we have

$$\forall \omega \in \widehat{O}: P(\omega) = \frac{1}{8}; \quad (49)$$

then, the variable distributions will be

$$P(\widehat{V}) = \left\{ V_1: \begin{cases} P(u) = \mathbf{0} \\ P(h_1) = \frac{1}{8} + \frac{1}{8} + \frac{1}{8} + \frac{1}{8} = \frac{\mathbf{1}}{\mathbf{2}} \\ P(t_1) = \frac{1}{8} + \frac{1}{8} + \frac{1}{8} + \frac{1}{8} = \frac{\mathbf{1}}{\mathbf{2}} \end{cases}, V_2: \begin{cases} P(u) = \mathbf{0} \\ P(h_2) = \frac{1}{8} + \frac{1}{8} + \frac{1}{8} + \frac{1}{8} = \frac{\mathbf{1}}{\mathbf{2}} \\ P(t_2) = \frac{1}{8} + \frac{1}{8} + \frac{1}{8} + \frac{1}{8} = \frac{\mathbf{1}}{\mathbf{2}} \end{cases} \right\}. \quad (50)$$

For $\widehat{\Theta}$, we can construct a conditional distribution $\Theta|\breve{\omega}$, where the known outcome is $\breve{\omega} = h_1$ (Formula 16)

$$\widehat{\Theta}|\breve{\omega} = \left\langle \begin{array}{c} \widehat{V}|\breve{\omega} = \{V_1: \{u, t_1\}, V_2: \{u, h_2, t_2\}\} \\ \widehat{R}|\breve{\omega} = \{r_1, r_2\} \\ \widehat{O}|\breve{\omega} = \{h_1 \cap {}^{r_1} h_2, h_1 \cap {}^{r_1} t_2, h_1 \cap {}^{r_2} h_2, h_1 \cap {}^{r_2} t_2\} \end{array} \right\rangle. \quad (51)$$

Then, we can also calculate the conditional PD of outcomes in $\widehat{\Theta}|\breve{\omega}$ (Formula 18):

$$P(\widehat{\Theta}|\breve{\omega}) = \begin{cases} P(h_1 \cap {}^{r_1} h_2) = \frac{P(h_1 \cap {}^{r_1} h_2)}{Z} = \frac{\frac{1}{8}}{\frac{1}{8} + \frac{1}{8} + \frac{1}{8} + \frac{1}{8}} = \frac{1}{4} \\ P(h_1 \cap {}^{r_1} t_2) = \cdots \\ P(h_1 \cap {}^{r_2} h_2) = \cdots \\ P(h_1 \cap {}^{r_2} t_2) = \cdots \end{cases}. \quad (52)$$

To make the example of a factorized network $\widetilde{\Theta}$ intuitive, we add one more coin to our experiment. Additionally, we add an extra variable $V_3$ to the model (Figures 6 and 7):

$$\widetilde{\Theta} = \begin{pmatrix} \widetilde{V} = [V_1:\{u, h_1, t_1\}, V_2:\{u, h_2, t_2\}, V_3:\{u, h_3, t_3\}] \\ \widetilde{R} = \{r_1, r_2\} \\ \widetilde{O} = [h_1 \cap^{r_1} h_2, t_1 \cap^{r_1} t_2, h_2 \cap^{r_1} h_3, t_2 \cap^{r_2} t_3] \end{pmatrix}. \quad (53)$$

In this network, we have two factors:

$$\Phi_1 = [h_1 \cap^{r_1} h_2, t_1 \cap^{r_1} t_2] \quad (54)$$

$$\Phi_2 = [h_2 \cap^{r_1} h_3, t_2 \cap^{r_2} t_3]. \quad (55)$$

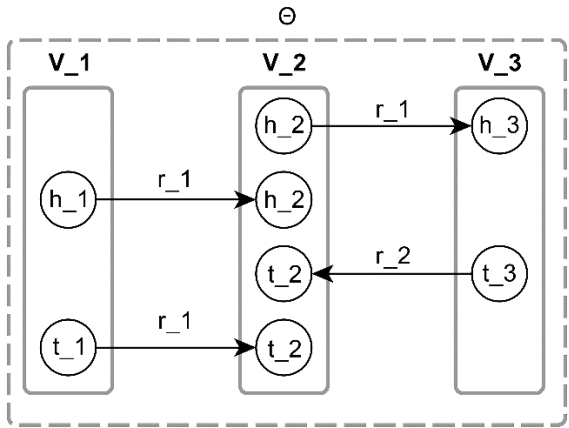

**Fig. 6** Outcomes of factorized network $\widetilde{\Theta}$

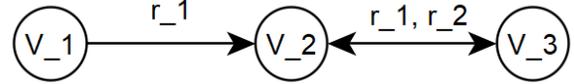

**Fig. 7** Factorized network $\widetilde{\Theta}$

## 3. Basic algorithms

The data structure described above will not be useful without algorithms to work with it. In this study, we explore some basic algorithms that implement the minimal necessary set of actions: building relation networks and probabilistic inference.

Notably, the algorithms described below are neither the most efficient nor the most interesting. They are added to this article merely to give the reader an intuition about what a PRN is. More advanced algorithms are topics for future works.

Next, we describe a relation network building algorithm and algorithms for computing conditional and full joint PDs. Furthermore, we present the practice part of the algorithms in Appendix C.

### 3.1. Building $\boldsymbol{O}$ by adding new outcomes to an algorithm

In the simplest case, network $\boldsymbol{O}$ can be built by manually defining the variables $\boldsymbol{V}$ and relation types $\boldsymbol{R}$ and adding outcomes $\omega$ to the set $\boldsymbol{O}$ as well as assigning them certain probabilities. Intuitively, you can think about it as wiring values from different variables $V$.

However, a more practical approach is to automatically build $\boldsymbol{O}$ from data. This is a key feature in developing a relation network model. Particularly, it is supposed to use online learning on a certain data stream. Similar approach was used in [14, 3].

The online learning algorithm might be as follows:

1. $\boldsymbol{O}: \{\omega \to \mathbb{N}\} = \emptyset$ # Define empty accumulator as a map from outcomes to counts
2. $\boldsymbol{T} = new\ DataSource$ # Let $T$ be some data source
3. $\forall\ t \in \boldsymbol{T}$: # For each experiment result
4. $\omega = outcome(t)$ # Build a new outcome base on input data
5. $if\ \omega \notin \boldsymbol{O}$: # If outcome is not in $\boldsymbol{O}$
6. $\boldsymbol{O} = \boldsymbol{O} \cup (\omega \to 0)$ # Then, add outcome
7. $\boldsymbol{O}(\omega) = \boldsymbol{O}(\omega) + 1$ # And increase its count
8. $\boldsymbol{O} = \begin{pmatrix} \boldsymbol{V} = \bigcup_{\omega \in \boldsymbol{O}} V(\omega) \\ \boldsymbol{R} = \bigcup_{\omega \in \boldsymbol{O}} r(\omega) \\ \boldsymbol{O} \end{pmatrix}$ # Compose result relation network

In this case, the total number of outcomes will be

$$N(\boldsymbol{\Theta}) = \sum_{\omega \in \boldsymbol{\Theta}.\boldsymbol{O}} \boldsymbol{\Theta}.\boldsymbol{O}(\omega) . (56)$$

The probability of some ω will be

$$P(\omega) = \frac{\boldsymbol{\Theta}.\boldsymbol{O}(\omega)}{N(\boldsymbol{\Theta})}, (57)$$

and the PD of all outcomes will be

$$P(\boldsymbol{\Theta}) = \left\{ \frac{\boldsymbol{\Theta}.\boldsymbol{O}(\omega)}{N(\boldsymbol{\Theta})} \mid \omega \in \boldsymbol{\Theta}.\boldsymbol{O} \right\}. (58)$$

### 3.2. Computation of conditional PD network $\boldsymbol{\Theta}|\breve{\omega}$ algorithm

An algorithm that builds $\boldsymbol{\Theta}|\breve{\omega}$ is quite simple. In fact, we just remove from the set $\boldsymbol{O}$ all outcomes that are known to be impossible, considering the evidence $\breve{\omega}$ (outcome reduction).

Algorithm for building $\boldsymbol{\Theta}|\breve{\omega}$:

1. $(\boldsymbol{\Theta}, \breve{\omega}) \rightarrow \boldsymbol{\Theta}|\breve{\omega}$:# Define a function that takes $\boldsymbol{\Theta}$ and $\breve{\omega}$ as input and returns $\boldsymbol{\Theta}|\breve{\omega}$
2. $\forall \omega \in \boldsymbol{\Theta}.\boldsymbol{O}$:# Scan over each outcome
3. $if\ (V(\breve{\omega}) \cap V(\omega) \neq \emptyset)$# If variables of $\breve{\omega}$ intersect with variables of $\omega$
4. $and\ (\dot{e}(\breve{\omega}) \nsubseteq \dot{e}(\omega))$:# And values of $\breve{\omega}$ is not a subset of values of $\omega$
5. $\boldsymbol{\Theta}.\boldsymbol{O} = \boldsymbol{\Theta}.\boldsymbol{O} - \omega$# Then, remove $\omega$
6. $return\ \boldsymbol{\Theta}|\breve{\omega}$# At the end, return processed $\boldsymbol{\Theta}$ as $\boldsymbol{\Theta}|\breve{\omega}$

In fact, the algorithm goes over all outcomes $\omega$ and removes those that intersect with $\breve{\omega}$ (have at least one value $e$ falling into the same variable $V$) and for which $\breve{\omega}$ is not a subgraph.

### 3.3. Computation of full joint PD network $\widetilde{\boldsymbol{\Theta}}$ algorithm

Obtaining a full joint PD $\widehat{\boldsymbol{\Theta}}$ on an arbitrary network, $\boldsymbol{\Theta}$ is impossible because such a network may not contain sufficient information about the probability dependences of all values $e$ from all variables $V$. Nevertheless, it is possible for the special case, a factorized network $\widetilde{\boldsymbol{\Theta}}$.

The network $\widetilde{\boldsymbol{\Theta}}$ is essentially a factor graph, and the algorithm below implements the factor product.

Algorithm for building $\widehat{\boldsymbol{\Theta}}$:

1. $\widetilde{\Theta} \to \widehat{\Theta}$:# Define a function that takes $\Theta$ as input and returns $\widehat{\Theta}$.
2. $\boldsymbol{\Phi} = \left\{\Phi_1, \Phi_2 \subseteq \widetilde{\boldsymbol{\Theta}}.\widetilde{\boldsymbol{O}} \,|\, V(\Phi_1) \cap V(\Phi_2) = V\big(e(\Phi_1) \cap e(\Phi_2)\big)\right\}$# Split outcomes on factors set
3. $\forall V \in \widetilde{\boldsymbol{\Theta}}.\widetilde{\boldsymbol{V}}$:# For each variable in $\widetilde{\Theta}$
4. $\Phi_V = \{\Phi \in \boldsymbol{\Phi} \,|\, V \in V(\Phi)\}$# Select a subset of factors that contains this variable
5. $\acute{\Phi} = \emptyset$# Create a new empty factor
6. $\forall e \in e(V)$:# For each value in variable V
7. $\forall \omega_1 \in \Phi_1(e), \dots, \forall \omega_n \in \Phi_n(e)$:# From $\Phi_V$ select $\omega$ that have $e$ and combine them
8. $if\ \forall V: E \in V(\omega_1) \cap \dots \cap V(\omega_n)$# If for each variable $V$ that in included in each of outcomes $\omega_1, \dots, \omega_n$
9. $\exists!\, e \in E, e(\omega_1), \dots, e(\omega_n)$:# Exist one $e$ which in $E$ and $\omega_1, \dots, \omega_n$
10. $\omega = \begin{pmatrix} e = \cup\, \omega_1.e, \dots, \omega_n.e \\ r = \cup\, \omega_1.r, \dots, \omega_n.r \end{pmatrix}$# Then, for each combination build a new joint outcome
11. $c = \prod \Phi_1(\omega_1), \dots, \Phi_n(\omega_n)$# And compute its count
12. $\acute{\Phi} = \acute{\Phi} \cup (\omega \to c)$# Then, add the joint outcome to $\acute{\Phi}$
13. $\boldsymbol{\Phi} = \boldsymbol{\Phi} - \Phi_V$# Remove selected factors $\Phi_V$ from $\boldsymbol{\Phi}$
14. $\boldsymbol{\Phi} = \boldsymbol{\Phi} \cup \acute{\Phi}$# And instead of $\Phi_V$, add the joint factor $\acute{\Phi}$
15. $return\ \widehat{\boldsymbol{\Theta}} = \begin{pmatrix} \widehat{\boldsymbol{V}} = \cup_{\Phi \in \boldsymbol{\Phi}}\, V(\Phi) \\ \widehat{\boldsymbol{R}} = \cup_{\Phi \in \boldsymbol{\Phi}}\, r(\Phi) \\ \widehat{\boldsymbol{O}} = \cup_{\Phi \in \boldsymbol{\Phi}}\, \Phi \end{pmatrix}$# When all variables are processed, return $\widehat{\Theta}$

Notably, the PD thus obtained is not the true one but is only a guess about what it might be. For example, if after conducting a series of experiments, we observe only outcomes $a \cap^r b$ and $b \cap^r c$, then we can assume that there is an outcome $a \cap^r b \cap^r c$ and build it using the above algorithm. Nevertheless, we will not know for sure whether the outcome $a \cap^r b \cap^r c$ is possible and what is its true probability until we begin to observe it.

### 3.4. Basic algorithms practice

We continue with the coin tossing experiment from the previous section, extending it to three coins, e.g., 1¢, 5¢, and 25¢.

After collecting data on coin tosses and applying the algorithm for constructing $\boldsymbol{\Theta}$ by adding new outcomes (see Appendix C), we obtained a network

$$\boldsymbol{\Theta} = \begin{pmatrix} \boldsymbol{V} = [V_1:\{u,t_1,h_1\}, V_2:\{u,h_2,t_2\}, V_3:\{u,h_3,t_3\}] \\ \boldsymbol{R} = \{r_1, r_2\} \\ \boldsymbol{O} = \begin{bmatrix} (h_1)^2, (h_2)^3, (t_3)^1, (h_2 \cap^{r_1} h_3)^1, \\ (h_1 \cap^{r_1} h_2)^1, (t_2 \cap^{r_2} t_3)^1, (t_1 \cap^{r_1} t_2)^1 \end{bmatrix} \end{pmatrix}. (59)$$

Additionally, the PD of all outcomes was calculated manually

$$P(\boldsymbol{\Theta}) = \left\{ \frac{\boldsymbol{\Theta}.\boldsymbol{O}(\omega)}{N(\boldsymbol{\Theta})} \mid \omega \in \boldsymbol{\Theta}.\boldsymbol{O} \right\} = \left\{ \frac{2}{10}, \frac{3}{10}, \frac{1}{10}, \frac{1}{10}, \frac{1}{10}, \frac{1}{10}, \frac{1}{10} \right\}. (60)$$

Using the same data and implementation in Python [15], we obtained the same result:

```
Outcomes:
ω = {(V2_t2)--{r2}--(V3_t3)}, c = 1, P(ω) = 0.1
ω = {(V2_h2)}, c = 3, P(ω) = 0.3
ω = {(V1_h1)}, c = 2, P(ω) = 0.2
ω = {(V2_h2)--{r1}--(V3_h3)}, c = 1, P(ω) = 0.1
ω = {(V3_t3)}, c = 1, P(ω) = 0.1
ω = {(V1_t1)--{r1}--(V2_t2)}, c = 1, P(ω) = 0.1
ω = {(V1_h1)--{r1}--(V2_h2)}, c = 1, P(ω) = 0.1
```

Applying the built $\boldsymbol{\Theta}$ the building $\boldsymbol{\Theta}|\breve{\omega}$ algorithm, we obtained a network containing a conditional PD

$$P(\boldsymbol{\Theta}|\breve{\omega}) = [(h_1 \cap^{r_1} h_2)^1, (t_3)^1]. (61)$$

Additionally, using the Python implementation [15], we obtained the PDs of outcomes

```
Outcomes:
ω = {(V1_h1)--{r1}--(V2_h2)}, c = 1, P(ω) = 0.5
ω = {(V3_t3)}, c = 1, P(ω) = 0.5
```

Because the built $\boldsymbol{\Theta}$ satisfies the factorized network $\widetilde{\boldsymbol{\Theta}}$ condition (Formulas 19 and 20), we could construct the full joint PD $\widehat{\boldsymbol{\Theta}}$ by applying the corresponding algorithm

$$\widehat{\boldsymbol{\Theta}} = \begin{pmatrix} \widehat{\boldsymbol{V}} = \bigcup_{\Phi \in \boldsymbol{\Phi}} V(\Phi) = [V_1:\{u,t_1,h_1\}, V_2:\{u,h_2,t_2\}, V_3:\{u,h_3,t_3\}] \\ \widehat{\boldsymbol{R}} = \bigcup_{\Phi \in \boldsymbol{\Phi}} r(\Phi) = \{r_1, r_2\} \\ \widehat{\boldsymbol{O}} = \bigcup_{\Phi \in \boldsymbol{\Phi}} \Phi = [(h_1 \cap^{r_1} h_2 \cap^{r_1} h_3)^6, (t_1 \cap^{r_1} t_2 \cap^{r_2} t_3)^1] \end{pmatrix}. (62)$$

We also obtained $\widehat{\boldsymbol{\Theta}}$ using the Python implementation [15]:

```
Outcomes:
ω = {(V1_h1)--{r1}--(V2_h2); (V2_h2)--{r1}--(V3_h3)}, c = 6, P(ω) = 0.8571428571428571
ω = {(V1_t1)--{r1}--(V2_t2); (V2_t2)--{r2}--(V3_t3)}, c = 1, P(ω) = 0.14285714285714285
```

## 4. Comparison with Bayesian and Markov networks

To improve the comprehension of the PRN, we compare it with two well-known probabilistic graphical models: Bayesian and Markov networks.

Because the PRN can contain a full joint PD for all variables included in it, we can model Bayesian and Markov networks on top of it.

Further, we show how to represent Bayes and Markov networks as a PRN and compare them numerically. Python implementation of the comparison is presented in Appendix D.

### 4.1. Representation of Bayesian and Markov networks as $\widetilde{\boldsymbol{\Theta}}$

We consider a Markov network [4] $G_m$ and Bayes network [16] $G_b$ represented as a factor graph [13]. Each factor $\phi(V_1, \ldots, V_n)$ can be converted into a set of outcomes $\Phi \subseteq \boldsymbol{O}$, which together form a factorized network $\widetilde{\boldsymbol{\Theta}}$.

For the Markov network, nonnormalized integer values included in $\phi(V_1: \boldsymbol{E}_1, \ldots, V_n: \boldsymbol{E}_n)$ can be directly converted into a set of outcomes:

$$\Phi = \{(e_1 \cap \ldots \cap e_n)^{\phi(e_1,\ldots,e_n)} \mid e_1 \in \boldsymbol{E}_1, \ldots, e_n \in \boldsymbol{E}_n\} \quad (63)$$

The factors of the Bayesian network are conveniently represented as conditional probability tables $P(V: \boldsymbol{E} \mid V_1: \boldsymbol{E}_1, \ldots, V_n: \boldsymbol{E}_n)$. They can also be directly converted into a set of outcomes:

$$\Phi = \{(e \cap^{V \leftarrow V_1,\ldots,V_n} e_1, \ldots, e_n)^{P(e|e_1,\ldots,e_n)*\eta} \mid e \in \boldsymbol{E}, e_1 \in \boldsymbol{E}_1, \ldots, e_n \in \boldsymbol{E}_n\}, \quad (64)$$

where $\eta$ denotes the conversion constant $\mathbb{R} * \eta = \mathbb{N}$.

Transforming each $\phi(V_1, \ldots, V_n)$ into its corresponding $\Phi$, we obtain a collection of sets $\boldsymbol{\Phi} \Leftrightarrow G_m$ and $\boldsymbol{\Phi} \Leftrightarrow G_b$, from which we can build a relation network:

$$\widetilde{\boldsymbol{\Theta}} = \left( \begin{array}{l} \widetilde{\boldsymbol{V}} = \bigcup_{\Phi \in \boldsymbol{\Phi}} V(\Phi) \\ \widetilde{\boldsymbol{R}} = \bigcup_{\Phi \in \boldsymbol{\Phi}} r(\Phi) \\ \widetilde{\boldsymbol{O}} = \bigcup_{\Phi \in \boldsymbol{\Phi}} \Phi \end{array} \right). \quad (65)$$

To the constructed network $\widetilde{\boldsymbol{\Theta}}$, we can apply the algorithm for calculating a conditional PD $\boldsymbol{O} \mid \widetilde{\omega}$ and algorithm for calculating a full joint PD $\widehat{\boldsymbol{\Theta}}$.

Combining the algorithms $\boldsymbol{O} \mid \widetilde{\omega}$ and $\widehat{\boldsymbol{\Theta}}$, we can perform the simplest inference [2]:

1. $(\widehat{\boldsymbol{\Theta}}, Y, X = e) \rightarrow P(Y|X = e)$:# Define a function that takes $\widehat{\boldsymbol{\Theta}}$, $Y$ and $E = e$ as input and returns $P(Y|E = e)$
2. $\forall\, e \in X$:# For each evidence value $e$
3. $\widehat{\boldsymbol{\Theta}} = \boldsymbol{\Theta}|\breve{\omega}\left(\widehat{\boldsymbol{\Theta}}, (e)\right)$# Calculate $\boldsymbol{\Theta}|\breve{\omega}$ where $e$ becomes a single vertex outcome $\breve{\omega}$
4. $\widehat{\boldsymbol{\Theta}} = \widehat{\boldsymbol{\Theta}}(\widehat{\boldsymbol{\Theta}})$# Then, calculate $\widehat{\boldsymbol{\Theta}}$
5. $return\ \{P(Y)|\ Y \in Y, \widehat{\boldsymbol{\Theta}}\}$# Return for each query variable $Y$ a calculated distribution $P(Y)$ based on $\widehat{\boldsymbol{\Theta}}$

### 4.2 Numerical comparison

For numerical comparison, we use two well-known and fair enough complex models: students' example (Figure 8) for the Bayesian network and friends' example (Figure 9) for the Markov network.

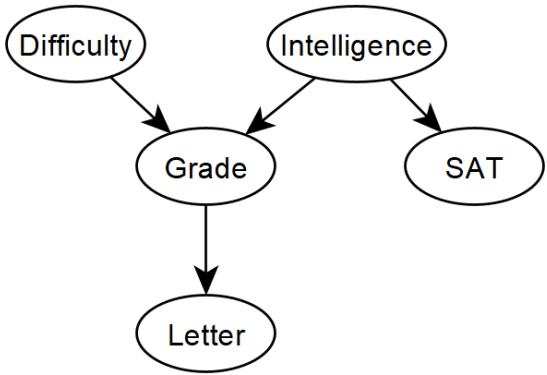
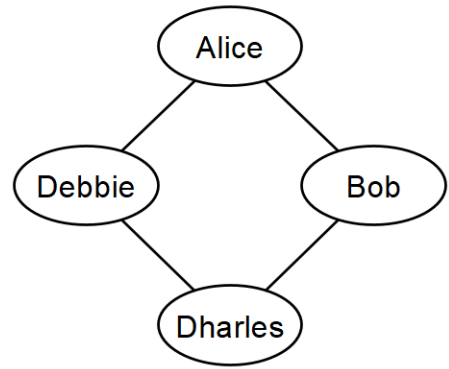

**Fig. 8** Students' Bayesian network

**Fig. 9** Friends' Markov network

As a reference implementation, we use Pgmpy [17] and compare it with the implementation of the PRN [15], for which we will write two scripts (see Appendix D): one for the Bayesian network and the other for the Markov network. Each script calculates the PD of the variables and the probabilistic inference for both implementations and then compares the results.

By running the first script for the Bayesian network, we obtain the following:

```
[comparing_joint_probability] At key = frozenset({'I(0)', 'S(0)', 'D(0)', 'G(0)', 'L(0)'})
factor_value = 0.011969999999999996, graph_value = 0.01197
```

```
[comparing_joint_probability] At key = frozenset({'S(1)', 'I(0)', 'D(0)', 'G(0)', 'L(0)'})
factor_value = 0.0006299999999999999, graph_value = 0.00063
[comparing_joint_probability] At key = frozenset({'I(0)', 'D(1)', 'S(0)', 'G(0)', 'L(0)'})
factor_value = 0.0013300000000000002, graph_value = 0.00133
[comparing_joint_probability] At key = frozenset({'S(1)', 'I(0)', 'D(1)', 'G(0)', 'L(0)'})
factor_value = 7.000000000000001e-05, graph_value = 7e-05
[comparing_joint_probability] At key = frozenset({'I(1)', 'S(0)', 'D(0)', 'G(0)', 'L(0)'})
factor_value = 0.0032400000000000007, graph_value = 0.00324
[comparing_joint_probability] At key = frozenset({'S(1)', 'I(1)', 'D(0)', 'G(0)', 'L(0)'})
factor_value = 0.012960000000000003, graph_value = 0.01296
[comparing_joint_probability] At key = frozenset({'I(1)', 'D(1)', 'S(0)', 'G(0)', 'L(0)'})
factor_value = 0.0012000000000000001, graph_value = 0.0012
[comparing_joint_probability] At key = frozenset({'S(1)', 'I(1)', 'D(1)', 'G(0)', 'L(0)'})
factor_value = 0.004800000000000004, graph_value = 0.0048
[comparing_joint_probability] At key = frozenset({'G(1)', 'I(0)', 'S(0)', 'D(0)', 'L(0)'})
factor_value = 0.06384000000000001, graph_value = 0.06384
[comparing_joint_probability] At key = frozenset({'G(1)', 'S(1)', 'I(0)', 'D(0)', 'L(0)'})
factor_value = 0.0033600000000000006, graph_value = 0.00336
[comparing_joint_probability] At key = frozenset({'G(1)', 'I(0)', 'S(0)', 'D(1)', 'L(0)'})
factor_value = 0.0266, graph_value = 0.0266
[comparing_joint_probability] At key = frozenset({'G(1)', 'S(1)', 'I(0)', 'D(1)', 'L(0)'})
factor_value = 0.0014, graph_value = 0.0014
[comparing_joint_probability] At key = frozenset({'G(1)', 'I(1)', 'S(0)', 'D(0)', 'L(0)'})
factor_value = 0.0011519999999999998, graph_value = 0.001152
[comparing_joint_probability] At key = frozenset({'G(1)', 'S(1)', 'I(1)', 'D(0)', 'L(0)'})
factor_value = 0.004607999999999999, graph_value = 0.004608
[comparing_joint_probability] At key = frozenset({'G(1)', 'I(1)', 'S(0)', 'D(1)', 'L(0)'})
factor_value = 0.00288, graph_value = 0.00288
[comparing_joint_probability] At key = frozenset({'G(1)', 'S(1)', 'I(1)', 'D(1)', 'L(0)'})
factor_value = 0.01152, graph_value = 0.01152
[comparing_joint_probability] At key = frozenset({'I(0)', 'G(2)', 'S(0)', 'D(0)', 'L(0)'})
factor_value = 0.11850299999999997, graph_value = 0.118503
[comparing_joint_probability] At key = frozenset({'S(1)', 'I(0)', 'G(2)', 'D(0)', 'L(0)'})
factor_value = 0.006236999999999995, graph_value = 0.006237
[comparing_joint_probability] At key = frozenset({'I(0)', 'G(2)', 'D(1)', 'S(0)', 'L(0)'})
factor_value = 0.18433799999999997, graph_value = 0.184338
[comparing_joint_probability] At key = frozenset({'S(1)', 'I(0)', 'G(2)', 'D(1)', 'L(0)'})
factor_value = 0.009702, graph_value = 0.009702
[comparing_joint_probability] At key = frozenset({'I(1)', 'G(2)', 'S(0)', 'D(0)', 'L(0)'})
factor_value = 0.0007128, graph_value = 0.0007128
[comparing_joint_probability] At key = frozenset({'S(1)', 'I(1)', 'G(2)', 'D(0)', 'L(0)'})
factor_value = 0.0028512, graph_value = 0.0028512
[comparing_joint_probability] At key = frozenset({'I(1)', 'G(2)', 'S(0)', 'D(1)', 'L(0)'})
factor_value = 0.004752000000000001, graph_value = 0.004752
[comparing_joint_probability] At key = frozenset({'S(1)', 'I(1)', 'G(2)', 'D(1)', 'L(0)'})
factor_value = 0.019008000000000004, graph_value = 0.019008
[comparing_joint_probability] At key = frozenset({'L(1)', 'I(0)', 'S(0)', 'D(0)', 'G(0)'})
factor_value = 0.10772999999999999, graph_value = 0.10773
[comparing_joint_probability] At key = frozenset({'L(1)', 'S(1)', 'I(0)', 'D(0)', 'G(0)'})
factor_value = 0.00567, graph_value = 0.00567
[comparing_joint_probability] At key = frozenset({'L(1)', 'I(0)', 'S(0)', 'D(1)', 'G(0)'})
factor_value = 0.011970000000000001, graph_value = 0.01197
[comparing_joint_probability] At key = frozenset({'L(1)', 'S(1)', 'I(0)', 'D(1)', 'G(0)'})
factor_value = 0.00063, graph_value = 0.00063
[comparing_joint_probability] At key = frozenset({'L(1)', 'I(1)', 'S(0)', 'D(0)', 'G(0)'})
factor_value = 0.029160000000000002, graph_value = 0.02916
[comparing_joint_probability] At key = frozenset({'L(1)', 'S(1)', 'I(1)', 'D(0)', 'G(0)'})
factor_value = 0.11664000000000001, graph_value = 0.11664
[comparing_joint_probability] At key = frozenset({'L(1)', 'I(1)', 'S(0)', 'D(1)', 'G(0)'})
factor_value = 0.010800000000000002, graph_value = 0.0108
[comparing_joint_probability] At key = frozenset({'L(1)', 'S(1)', 'I(1)', 'D(1)', 'G(0)'})
factor_value = 0.04320000000000001, graph_value = 0.0432
[comparing_joint_probability] At key = frozenset({'G(1)', 'L(1)', 'I(0)', 'S(0)', 'D(0)'})
factor_value = 0.09575999999999997, graph_value = 0.09576
[comparing_joint_probability] At key = frozenset({'G(1)', 'L(1)', 'S(1)', 'I(0)', 'D(0)'})
factor_value = 0.005039999999999999, graph_value = 0.00504
[comparing_joint_probability] At key = frozenset({'G(1)', 'L(1)', 'I(0)', 'S(0)', 'D(1)'})
factor_value = 0.0399, graph_value = 0.0399
```

```
[comparing_joint_probability] At key = frozenset({'G(1)', 'L(1)', 'S(1)', 'I(0)', 'D(1)'})
factor_value = 0.0021000000000000003, graph_value = 0.0021
[comparing_joint_probability] At key = frozenset({'G(1)', 'L(1)', 'I(1)', 'S(0)', 'D(0)'})
factor_value = 0.0017280000000000002, graph_value = 0.001728
[comparing_joint_probability] At key = frozenset({'G(1)', 'L(1)', 'S(1)', 'I(1)', 'D(0)'})
factor_value = 0.006912000000000001, graph_value = 0.006912
[comparing_joint_probability] At key = frozenset({'G(1)', 'L(1)', 'I(1)', 'S(0)', 'D(1)'})
factor_value = 0.00432, graph_value = 0.00432
[comparing_joint_probability] At key = frozenset({'G(1)', 'L(1)', 'S(1)', 'I(1)', 'D(1)'})
factor_value = 0.01728, graph_value = 0.01728
[comparing_joint_probability] At key = frozenset({'L(1)', 'I(0)', 'G(2)', 'S(0)', 'D(0)'})
factor_value = 0.001197, graph_value = 0.001197
[comparing_joint_probability] At key = frozenset({'L(1)', 'S(1)', 'I(0)', 'G(2)', 'D(0)'})
factor_value = 6.3e-05, graph_value = 6.3e-05
[comparing_joint_probability] At key = frozenset({'L(1)', 'I(0)', 'G(2)', 'S(0)', 'D(1)'})
factor_value = 0.0018619999999999995, graph_value = 0.001862
[comparing_joint_probability] At key = frozenset({'L(1)', 'S(1)', 'I(0)', 'G(2)', 'D(1)'})
factor_value = 9.799999999999998e-05, graph_value = 9.8e-05
[comparing_joint_probability] At key = frozenset({'L(1)', 'I(1)', 'G(2)', 'S(0)', 'D(0)'})
factor_value = 7.2e-06, graph_value = 7.2e-06
[comparing_joint_probability] At key = frozenset({'L(1)', 'S(1)', 'I(1)', 'G(2)', 'D(0)'})
factor_value = 2.88e-05, graph_value = 2.88e-05
[comparing_joint_probability] At key = frozenset({'L(1)', 'I(1)', 'G(2)', 'S(0)', 'D(1)'})
factor_value = 4.8e-05, graph_value = 4.8e-05
[comparing_joint_probability] At key = frozenset({'L(1)', 'S(1)', 'I(1)', 'G(2)', 'D(1)'})
factor_value = 0.000192, graph_value = 0.000192
[comparing_inference] margin_factor_i = {'L': {'L(0)': 0.6113999999999999, 'L(1)':
0.38859999999999995}, 'G': {'G(0)': 0.2, 'G(1)': 0.33999999999999997, 'G(2)': 0.4599999999999999},
'D': {'D(0)': 0.6000000000000001, 'D(1)': 0.39999999999999997}, 'S': {'S(0)': 0.95, 'S(1)': 0.05}}
[comparing_inference] margin_graph_i = {'I': {'I(0)': 1.0}, 'D': {'D(1)': 0.4, 'D(0)': 0.6}, 'G':
{'G(1)': 0.34, 'G(2)': 0.46, 'G(0)': 0.2}, 'L': {'L(0)': 0.6114, 'L(1)': 0.3886}, 'S': {'S(1)': 0.05,
'S(0)': 0.95}}
[comparing_inference] margin_factor_ig = {'L': {'L(0)': 0.09999999999999998, 'L(1)':
0.8999999999999999}, 'S': {'S(0)': 0.95, 'S(1)': 0.05}, 'D': {'D(0)': 0.8999999999999999, 'D(1)':
0.1}}
[comparing_inference] margin_graph_ig = {'I': {'I(0)': 1.0}, 'D': {'D(1)': 0.1, 'D(0)': 0.9}, 'G':
{'G(0)': 1.0}, 'L': {'L(0)': 0.1, 'L(1)': 0.9}, 'S': {'S(0)': 0.05, 'S(0)': 0.95}}
```

By running the second script for the Markov network, we obtain the following:

```
Joint markov prop: ['P(A_0, B_0, C_0, D_0) = 0.0416560212390167', 'P(A_0, B_0, C_0, D_1) =
0.0416560212390167', 'P(A_0, B_0, C_1, D_0) = 0.0416560212390167', 'P(A_0, B_0, C_1, D_1) =
4.16560212390167e-06', 'P(A_0, B_1, C_0, D_0) = 6.942670206502783e-05', 'P(A_0, B_1, C_0, D_1) =
6.942670206502783e-05', 'P(A_0, B_1, C_1, D_0) = 0.6942670206502782', 'P(A_0, B_1, C_1, D_1) =
6.942670206502783e-05', 'P(A_1, B_0, C_0, D_0) = 1.3885340413005566e-05', 'P(A_1, B_0, C_0, D_1) =
0.13885340413005565', 'P(A_1, B_0, C_1, D_0) = 1.3885340413005566e-05', 'P(A_1, B_0, C_1, D_1) =
1.3885340413005566e-05', 'P(A_1, B_1, C_0, D_0) = 1.3885340413005566e-06', 'P(A_1, B_1, C_0, D_1) =
0.013885340413005565', 'P(A_1, B_1, C_1, D_0) = 0.013885340413005565', 'P(A_1, B_1, C_1, D_1) =
0.013885340413005565']
Joint relation graph prop: ['P(A_0, B_0, C_0, D_0) = 0.0416560212390167', 'P(A_0, B_0, C_0, D_1) =
0.0416560212390167', 'P(A_0, B_0, C_1, D_0) = 0.0416560212390167', 'P(A_0, B_0, C_1, D_1) =
4.16560212390167e-06', 'P(A_0, B_1, C_0, D_0) = 6.942670206502783e-05', 'P(A_0, B_1, C_0, D_1) =
6.942670206502783e-05', 'P(A_0, B_1, C_1, D_0) = 0.6942670206502782', 'P(A_0, B_1, C_1, D_1) =
6.942670206502783e-05', 'P(A_1, B_0, C_0, D_0) = 1.3885340413005566e-05', 'P(A_1, B_0, C_0, D_1) =
0.13885340413005565', 'P(A_1, B_0, C_1, D_0) = 1.3885340413005566e-05', 'P(A_1, B_0, C_1, D_1) =
1.3885340413005566e-05', 'P(A_1, B_1, C_0, D_0) = 1.3885340413005566e-06', 'P(A_1, B_1, C_0, D_1) =
0.013885340413005565', 'P(A_1, B_1, C_1, D_0) = 0.013885340413005565', 'P(A_1, B_1, C_1, D_1) =
0.013885340413005565']
Inference E=A_0 markov prop = ['P(A_0, B_0, C_0, D_0) = 0.050834275179487354', 'P(A_0, B_0, C_0, D_1)
= 0.050834275179487354', 'P(A_0, B_0, C_1, D_0) = 0.050834275179487354', 'P(A_0, B_0, C_1, D_1) =
5.0834275179487354e-06', 'P(A_0, B_1, C_0, D_0) = 8.472379196581226e-05', 'P(A_0, B_1, C_0, D_1) =
8.472379196581226e-05', 'P(A_0, B_1, C_1, D_0) = 0.8472379196581226', 'P(A_0, B_1, C_1, D_1) =
8.472379196581226e-05']
Inference E=A_0 relation graph prop = ['P(A_0, B_0, C_0, D_0) = 0.050834275179487354', 'P(A_0, B_0,
C_0, D_1) = 0.050834275179487354', 'P(A_0, B_0, C_1, D_0) = 0.050834275179487354', 'P(A_0, B_0, C_1,
D_1) = 5.0834275179487354e-06', 'P(A_0, B_1, C_0, D_0) = 8.472379196581226e-05', 'P(A_0, B_1, C_0,
D_1) = 8.472379196581226e-05', 'P(A_0, B_1, C_1, D_0) = 0.8472379196581226', 'P(A_0, B_1, C_1, D_1) =
8.472379196581226e-05']
```

## 5. Conclusion

In this work, we presented a novel approach to model probabilistic relationships of arbitrary type between random variables. We started with a simple idea and developed it to be a formal definition of PRN $\boldsymbol{\Theta}$. Next, we explored basic model properties, such as full-joint PD $\widehat{\boldsymbol{\Theta}}$ and conditional PD $\boldsymbol{\Theta}|\widetilde{\omega}$. In addition, we considered a special case of PRN—factorized relation graph $\widetilde{\widehat{\boldsymbol{\Theta}}}$, which will be useful in modeling Bayesian and Markov networks.

A given probabilistic graphical model will be useless for real-life applications, if it did not have algorithms for its construction and probabilistic inference. In this article, we talked about the simplest versions of such algorithms. Furthermore, we proposed an algorithm for constructing a full-joint PD for the factorized relation graph $\widetilde{\widehat{\boldsymbol{\Theta}}}$. Although this algorithm demonstrated the fundamental possibility of building a full-joint PD, it will unlikely be used in real-life applications.

To show the validity and practical utility of the proposed approach, we implemented PRN and its algorithms in Python [15]. Additionally, we performed a numerical comparison of our implementation with the known implementations of Bayesian and Markov networks [17]. As examples, we have chosen well-known and simple models, Students' Bayesian network, and Friends' Markov network for comparison. For each of these models, we calculated the full-joint and conditional PDs, and it match exactly for all implementations.

Initially, the PRN was created for a purely practical purpose, that is, to serve as a kernel of large expert and decision-making systems. Although it is too early to talk about real commercial applications at this stage, the model can be useful as a simple means to show or visualize and help to understand the joint distribution of random variables. Moreover, PRN can be a foundation for developing better learning and inference algorithms and be a starting point for the future research.

## 6. Future work

A significant drawback of the current software implementation [15] is its representation as an array of outcome graphs, which is inconvenient for organizing calculations. Collapsing the outcome graphs into a "flat"-like structure, such as Gibbs distribution, would allow to considerably reduce the amount of machine memory occupied and parallelize the calculations of different segments of the model. For example, an interesting approach that can be used here is shown in a previous study [14]. Generally, the development of a memory-efficient digital representation of the model is an important future task.

In this study, we describe an algorithm for calculating a full joint PD $\widehat{\Theta}$ for a factorized network $\widetilde{\Theta}$. It is obvious that the calculation of $\widetilde{\Theta}$ for any complex model will require huge computational resources and hence is unadvisable in practice. The good news is that often in practice, we do not need $\widehat{\Theta}$, but it is sufficient to build just some local joint distribution for variables of interest, by manipulating a small part of the model. For example, if the model has a pair of outcomes a $\cap^r$ b and b $\cap^r$ c, using some heuristics, our algorithm can assume that they are interconnected and build for them a joint distribution a $\cap^r$ b $\cap^r$ c, which will be used in the probabilistic inference. Thus, developing an efficient algorithm for computing local joint distribution can be interesting in the future.

Another interesting topic for research in the future is applying continuous random variables in the PRN. Theoretically, we can think of them as having an infinite number of values, so the model must have an infinite number of outcomes, which, of course, is infeasible in practice. The idea here is to replace the set of edges connecting the values of variables with a continuous bijective function. Thus, all or part of the variables of the model, as it were, will be connected by functions.

**Appendix A – Theorem of the sum of probabilities of variable values**

Let $(\Omega, \mathcal{F}, P)$ be a probability space in which the probability of each outcome is $P(\omega) = \lim_{|T|\to\infty} \frac{|\omega(T)|}{|T|}$, and $V: \Omega \to E$ is a random variable defined on the space $(\Omega, \mathcal{F}, P)$, so $\boldsymbol{E} = \mathcal{F}$ and the probability of each value $P(e) = \begin{cases} \sum_{\{\omega \in \Omega | e(\omega) \ni e\}} P(\omega); & e \neq u \\ 1 - \sum_{\{e \in E | e \neq u\}} P(e); & e = u \end{cases}$. Then, the sum of the probabilities of all values of $V$ wil=l be $\sum_{e \in E} P(e) = 1$.

Proof:

- The sum of the probabilities of all outcomes is always $\sum_{\omega \in \Omega} P(\omega) = 1$
- Each outcome $\omega \in \Omega$ can contain only one $e \in \boldsymbol{E}$: $\forall e_1 \in \boldsymbol{E}, e_2 \in \boldsymbol{E}, \omega \in \Omega$: $\nexists\, e(\omega) \ni e_1 \cap e_2, e_1 \neq e_2$
- Therefore, the sets of outcomes containing $e \in \boldsymbol{E}$ do not intersect: $\bigcap_{e \in E;\ e \neq u} \{\omega \in \Omega | e(\omega) \ni e\} = \emptyset$
- Thus, the sum of their probabilities: $\sum_{e \in E;\ e \neq u} \sum_{\{\omega \in \Omega | e(\omega) \ni e\}} P(\omega) \leq 1$
- Additionally, the sum of the probabilities of the values $e \in \boldsymbol{E}$, except for the value $u$: $\sum_{e \in E;\ e \neq u} P(e) \leq 1$
- Considering the case $e = u$, we obtain $\sum_{e \in E} P(e) = 1 - \sum_{e \in E;\ e \neq u} P(e) + \sum_{e \in E;\ e \neq u} P(e) = 1$
- Q.E.D.

## Appendix B – Theorem of condition independency for $\widehat{\boldsymbol{\Theta}}$

Let $\widehat{\boldsymbol{\Theta}} = \langle \widehat{\boldsymbol{V}}, \widehat{\boldsymbol{R}}, \widehat{\boldsymbol{O}} \rangle$, a factorized PRN comprising a set of variables $\widehat{\boldsymbol{V}}$ and a set of outcomes $\widehat{\boldsymbol{R}}$; in this case, two arbitrary vertices $A, B \in \widehat{\boldsymbol{V}}$ are conditionally independent (i.e., $A \perp\!\!\!\perp B | Z$) if and only if all possible paths $A \cap \ldots \cap B$ are blocked by observable vertices $Z \in \boldsymbol{Z} \subseteq \widehat{\boldsymbol{V}}$ (i.e., $A \cap \ldots \cap Z \cap \ldots \cap B$) and outcome set $\widehat{\boldsymbol{O}}$ does not contain an outcome $\omega$ such that it simultaneously includes the value $z$ and its neighboring values lying on the path $A \cap \ldots \cap B$ (e.g., $a \cap \ldots \cap c \cap z \cap d \cap \ldots \cap b$).

Proof:

- Consider the simplest network $\boldsymbol{\Theta}$ of three variables—$A$, $B$, $Z$—and of two factors—$\Phi_{A,Z}$, $\Phi_{Z,B}$. The graph of which looks like $A - Z - B$.
- For the case $\Phi_{A,Z} = \emptyset$ or $\Phi_{Z,B} = \emptyset$, the product of the factors $\Phi_{A,Z} * \Phi_{Z,B} = \emptyset$. Therefore, $A \perp\!\!\!\perp B$, $A \perp\!\!\!\perp Z$, $B \perp\!\!\!\perp Z$.
- For the case $\Phi_{A,Z} \neq \emptyset$ and $\Phi_{Z,B} \neq \emptyset$ and $Z = z$, all outcomes that do not include the observed value of $Z = z$ will be removed from the factors $\Phi_{A,Z}|Z = \{\omega \in \Phi_{A,Z} | z \in e(\omega)\}$; then, when the factors $\Phi_{A,Z}|Z * \Phi_{Z,B}|Z$ are multiplied, the probability of each of the outcomes $\omega_{A,Z} \in \Phi_{A,Z}|Z$ will be multiplied by the probability of each of the outcomes $\omega_{Z,B} \in \Phi_{Z,B}|Z$. Thus, the probability of each value of the variables $A$ will be multiplied by the sum of the probabilities of the values of $B$ (i.e., by the same number), and vice versa. Therefore, changing the PD of values from $A$ will not change the PD from $B$. Consequently, $A \perp\!\!\!\perp B | Z$. Notably, this statement will not be true for the factor $\Phi_{A,Z,B}$, i.e., the factor containing the outcomes $a \cap z \cap b$ that simultaneously include values from all three variables.
- Variables added to the beginning or end of the $X - \cdots - A - Z - B - \cdots - Y$ chain with $Z = z$ will also be probabilistically independent $X - \cdots - A \perp\!\!\!\perp B - \cdots - Y | Z$ because by multiplying the factors $\Phi_{X,\ldots} * \ldots * \Phi_{\ldots,A} * \Phi_{A,Z} = \Phi_{X,\ldots,Z}$ and $\Phi_{Z,B} * \Phi_{B,\ldots} * \ldots * \Phi_{\ldots,Y} = \Phi_{Z,\ldots,Y}$, we still get a case with two factors $\Phi_{X,\ldots,Z}$ and $\Phi_{Z,\ldots,Y}$ similar to the previous one.
- Having several paths $A - Z_1 \ldots Z_n - B$ between $A$ and $B$, with $Z_1 = z_1, \ldots, Z_n = z_n$, the same as for a single path, the outcomes of unobserved values will be removed from the factors $\genfrac{}{}{0pt}{}{\Phi_{A,Z_1}|Z_1=\{\omega\in\Phi_{A,Z_1}|z_1\in e(\omega)\}}{\Phi_{A,Z_n}|Z_n=\{\omega\in\Phi_{A,Z_n}|z_n\in e(\omega)\}}$ and $\genfrac{}{}{0pt}{}{\Phi_{Z_1,B}|Z_1=\{\omega\in\Phi_{Z_1,B}|z_2\in e(\omega)\}}{\Phi_{Z_n,B}|Z_n=\{\omega\in\Phi_{Z_n,B}|z_n\in e(\omega)\}}$; then, in each product $\Phi_{A,Z_1}|Z_1 * \Phi_{Z_1,B}|Z_1, \dots, \Phi_{A,Z_n}|Z_n * \Phi_{Z_n,B}|Z_n$, the probability of each value of variable $A$ will be multiplied by the sum of the probabilities of the values of $B$ (i.e., by the same number), and vice versa, which is the same as in the case of one path and will not change the PD from $B$ when it changes for $A$, and vice versa. Therefore, $A \perp\!\!\!\perp B | Z_1 \dots Z_n$.

- Q.E.D.

**Appendix C – Basic practice algorithms**

### C.1. Building a $\boldsymbol{\Theta}$ algorithm by adding new outcomes:

- Extend the coin toss experiment (from Section 2) to three coins, e.g., 1¢, 5¢, and 25¢.
- Imagine that we are watching a person tossing coins. Each toss is an experiment. He/she squeezes some coins (possibly none) of different denominations in his/her fist and then opens his/her fist, so the coins fall to the floor, and we write down the observed result of the experiment: what denominations the coins were, how they fell (head, tail), and in what sequence.
- Let us represent such an experiment as a model of three random variables $V_1, V_2, V_3$, each of which corresponds to the denomination of a coin and the values of the variables $u, t, h$ correspond to the experiment's outcome.
- Suppose we performed 10 experiments and got a set of observations (see Table 1 column $\boldsymbol{T}$).
- Using a simple function $outcome(t)$, we can build a set of observed outcomes, one $\omega$ for each observation (see Table 1 column $\omega$).
- As outcomes are analyzed, we accumulate them in $\boldsymbol{O}$ (see Table 1 columns $\boldsymbol{O}$).
- Now, from the accumulated outcomes, we can construct $\boldsymbol{\Theta}$:

$$\boldsymbol{\Theta} = \begin{pmatrix} \boldsymbol{V} = [V_1:\{u, t_1, h_1\}, V_2:\{u, h_2, t_2\}, V_3:\{u, h_3, t_3\}] \\ \boldsymbol{R} = \{r_1, r_2\} \\ \boldsymbol{O} = \begin{bmatrix} (h_1)^2, (h_2)^3, (t_3)^1, (h_2 \cap^{r_1} h_3)^1 \\ (h_1 \cap^{r_1} h_2)^1, (t_2 \cap^{r_2} t_3)^1, (t_1 \cap^{r_1} t_2)^1 \end{bmatrix} \end{pmatrix}$$

- Graphically, the resulting network can be represented as in Figure 10.
- Next, we can count the number of outcomes:

$$N(\boldsymbol{\Theta}) = \sum_{\omega \in \boldsymbol{\Theta}.\boldsymbol{O}} \boldsymbol{\Theta}.\boldsymbol{O}(\omega) = 1 + 2 + 2 + 1 + 1 + 1 + 1 + 1 = 10$$

- The probability of each outcome can be calculated, e.g.:

$$P(h_1) = \frac{\boldsymbol{\Theta}.\boldsymbol{O}(\omega)}{N(\boldsymbol{\Theta})} = \frac{2}{10}$$

- In addition, the PD of all outcomes can be obtained:

$$P(\boldsymbol{\Theta}) = \left\{ \frac{\boldsymbol{\Theta}.\boldsymbol{O}(\omega)}{N(\boldsymbol{\Theta})} \mid \omega \in \boldsymbol{\Theta}.\boldsymbol{O} \right\} = \left\{ \frac{2}{10}, \frac{3}{10}, \frac{1}{10}, \frac{1}{10}, \frac{1}{10}, \frac{1}{10}, \frac{1}{10} \right\}$$

- Python example implementation: building_net_from_coin_toss.py

## C.2. Computation of conditional PD network $\boldsymbol{\Theta}|\breve{\omega}$ algorithm:

- Continue with the built relation network (Figure 10):

  $$\boldsymbol{\Theta}.\boldsymbol{O} = [(h_1)^2, (h_2)^3, (t_3)^1, (h_2 \cap^{r_1} h_3)^1, (h_1 \cap^{r_1} h_2)^1, (t_2 \cap^{r_2} t_3)^1, (t_1 \cap^{r_1} t_2)^1]$$

- Suppose we observe evidence $\breve{\omega} = h_1 \cap^{r_1} h_2$, in this case, we remove all outcomes that contain values from $V_1$ or $V_2$, except those that contain $h_1 \cap^{r_1} h_2$ as a subgraph:

  $$P(\boldsymbol{\Theta}|\breve{\omega}) = [(h_1 \cap^{r_1} h_2)^1, (t_3)^1]$$

- Graphically, the resulting network can be represented as in Figure 11.
- Next, we can calculate the probability of each outcome, e.g.:

  $$P(t_3) = \frac{\boldsymbol{\Theta}.c(t_3)}{n} = \frac{1}{2}$$

- Python example implementation: building_net_with_conditional_probability.py

| T | ω | O |
|---|---|---|
| 5¢ H | $\omega_1 = h_2$ | $[(h_2)^1]$ |
| 1¢ H | $\omega_2 = h_1$ | $[(h_2)^1, (h_1)^1]$ |
| 5¢ H | $\omega_3 = h_2$ | $[(h_2)^2, (h_1)^1]$ |
| 25¢ T | $\omega_4 = t_3$ | $[(h_2)^2, (h_1)^1, (t_3)^1]$ |
| 1¢ H | $\omega_5 = h_1$ | $[(h_2)^2, (h_1)^2, (t_3)^1]$ |
| 5¢ H | $\omega_6 = h_2$ | $[(h_2)^3, (h_1)^2, (t_3)^1]$ |
| 5¢ H then 25¢ H | $\omega_7 = h_2 \cap^{r_1} h_3$ | $[(h_2)^3, (h_1)^2, (t_3)^1, (h_2 \cap^{r_1} h_3)^1]$ |
| 1¢ H then 5¢ H | $\omega_8 = h_1 \cap^{r_1} h_2$ | $[(h_2)^3, (h_1)^2, (t_3)^1, (h_2 \cap^{r_1} h_3)^1, (h_1 \cap^{r_1} h_2)^1]$ |
| 25¢ T then 5¢ T | $\omega_9 = t_2 \cap^{r_2} t_3$ | $[(h_2)^3, (h_1)^2, (t_3)^1, (h_2 \cap^{r_1} h_3)^1, (h_1 \cap^{r_1} h_2)^1, (t_2 \cap^{r_2} t_3)^1]$ |
| 1¢ T then 5¢ T | $\omega_{10} = t_1 \cap^{r_1} t_2$ | $[(h_2)^3, (h_1)^2, (t_3)^1, (h_2 \cap^{r_1} h_3)^1, (h_1 \cap^{r_1} h_2)^1, (t_2 \cap^{r_2} t_3)^1, (t_1 \cap^{r_1} t_2)^1]$ |

*Table 1. Two-coin-toss experiment*

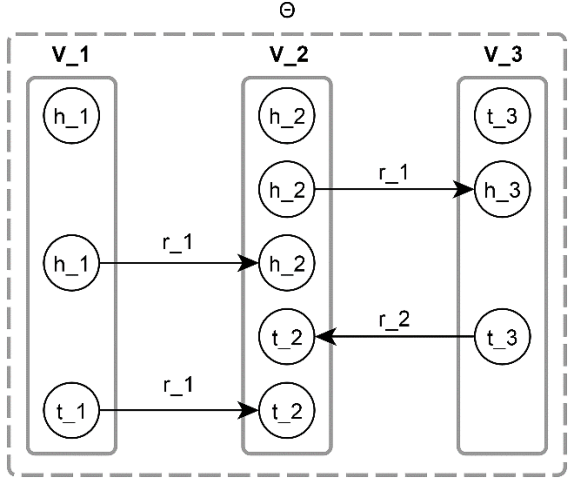

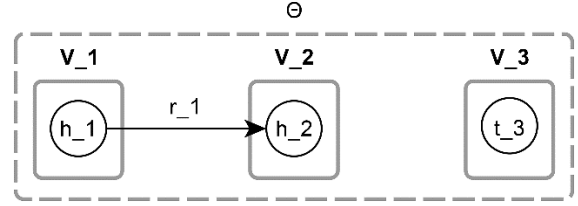

Fig. 10 Built network $\Theta$

Fig. 11 $\Theta|\widecheck{\omega}$ network

### C.3. Computation of joint distribution network $\widehat{\Theta}$ algorithm:

- The network $\Theta$ we got satisfies the condition of a factorized network $\widetilde{\Theta}$ (Formula 14), so we can use it to construct a network $\widehat{\Theta}$ containing the full joint PD.
- Let us split the set of outcomes of the network $\widetilde{\Theta}$:

$$\widetilde{\Theta} = [(h_1)^2, (h_2)^3, (t_3)^1, (h_2 \cap^{r_1} h_3)^1, (h_1 \cap^{r_1} h_2)^1, (t_2 \cap^{r_2} t_3)^1, (t_1 \cap^{r_1} t_2)^1]$$

On the set of factors,

$$\Phi = \begin{cases} \Phi_1 = [(h_1)^2] \\ \Phi_2 = [(h_1 \cap^{r_1} h_2)^1, (t_1 \cap^{r_1} t_2)^1] \\ \Phi_3 = [(h_2)^3] \\ \Phi_4 = [(h_2 \cap^{r_1} h_3)^1, (t_2 \cap^{r_2} t_3)^1] \\ \Phi_5 = [(t_3)^1] \end{cases}$$

- For each network variable from $\widetilde{V}$,
  - $V_1$:
    - $\Phi_V = \begin{cases} \Phi_1 = [(h_1)^2] \\ \Phi_2 = [(h \cap^{r_1} h_2)^1, (t_1 \cap^{r_1} t_2)^1] \end{cases}$
    - For each value from the $V_1$ domain,
      - $h_1: \omega_1 = (h_1)^2 * (h_1 \cap^{r_1} h_2)^1 = (h_1 \cap^{r_1} h_2)^2$

- $t_1$: $\omega_2 = (t_1 \cap {}^{r_1} t_2)^1$

- $\Phi = \Phi - \Phi_V \cup \{\omega_1, \omega_2\} = \begin{cases} \Phi_3 = [(h_2)^3] \\ \Phi_4 = [(h_2 \cap {}^{r_1} h_3)^1, (t_2 \cap {}^{r_2} t_3)^1] \\ \Phi_5 = [(t_3)^1] \\ \Phi_6 = [(h_1 \cap {}^{r_1} h_2)^2, (t_1 \cap {}^{r_1} t_2)^1] \end{cases}$

- $V_2$:
  - $\Phi_V = \begin{cases} \Phi_3 = [(h_2)^3] \\ \Phi_4 = [(h_2 \cap {}^{r_1} h_3)^1, (t_2 \cap {}^{r_2} t_3)^1] \\ \Phi_6 = [(h_1 \cap {}^{r_1} h_2)^2, (t_1 \cap {}^{r_1} t_2)^1] \end{cases}$
  - For each value from the $V_1$ domain,
    - $h_1$: $\omega_1 = (h_1 \cap {}^{r_1} h_2)^2 * (h_2)^3 * (h_2 \cap {}^{r_1} h_3)^1 = (h_1 \cap {}^{r_1} h_2 \cap {}^{r_1} h_3)^6$
    - $t_1$: $\omega_2 = (t_1 \cap {}^{r_1} t_2)^1 * (t_2 \cap {}^{r_2} t_3)^1 = (t_1 \cap {}^{r_1} t_2 \cap {}^{r_2} t_3)^1$
  - $\Phi = \Phi - \Phi_V \cup \{\omega_1, \omega_2\} =$
    $\begin{cases} \Phi_5 = [(t_3)^1] \\ \Phi_7 = [(h_1 \cap {}^{r_1} h_2 \cap {}^{r_1} h_3)^6, (t_1 \cap {}^{r_1} t_2 \cap {}^{r_2} t_3)^1] \end{cases}$

- $V_3$:
  - $\Phi_V = \begin{cases} \Phi_5 = [(t_3)^1] \\ \Phi_7 = [(h_1 \cap {}^{r_1} h_2 \cap {}^{r_1} h_3)^6, (t_1 \cap {}^{r_1} t_2 \cap {}^{r_2} t_3)^1] \end{cases}$
  - For each value from the $V_1$ domain,
    - $h_1$: $\omega_1 = (h_1 \cap {}^{r_1} h_2 \cap {}^{r_1} h_3)^6$
    - $t_1$: $\omega_2 = (t_1 \cap {}^{r_1} t_2 \cap {}^{r_2} t_3)^1 * (t_3)^1 = (t_1 \cap {}^{r_1} t_2 \cap {}^{r_2} t_3)^1$
  - $\Phi = \Phi - \Phi_V \cup \{\omega_1, \omega_2\} = \{\Phi_7 = [(h_1 \cap {}^{r_1} h_2 \cap {}^{r_1} h_3)^6, (t_1 \cap {}^{r_1} t_2 \cap {}^{r_2} t_3)^1]\}$

o Thus, we obtain

$$\widehat{\boldsymbol{\theta}} = \begin{pmatrix} \widehat{V} = \bigcup_{\Phi \in \Phi} V(\Phi) = [V_1: \{u, t_1, h_1\}, V_2: \{u, h_2, t_2\}, V_3: \{u, h_3, t_3\}] \\ \widehat{R} = \bigcup_{\Phi \in \Phi} r(\Phi) = \{r_1, r_2\} \\ \widehat{O} = \bigcup_{\Phi \in \Phi} \Phi = [(h_1 \cap {}^{r_1} h_2 \cap {}^{r_1} h_3)^6, (t_1 \cap {}^{r_1} t_2 \cap {}^{r_2} t_3)^1] \end{pmatrix}$$

o Python example implementation: building_net_with_joint_probability.py

# Appendix D - Numerical comparison with Bayesian and Markov networks

## D.1. Comparing with the Bayesian network (comparing_with_bayes_network.py)

```python
net_config = {
    'D': (2, [[.6], [.4]], None, None),   # Difficulty
    'I': (2, [[.7], [.3]], None, None),   # Intelligence
    'G': (3, [
        [.3, .05, .9, .5],
        [.4, .25, .08, .3],
        [.3, .7, .02, .2]], ['I', 'D'], [2, 2]),   # Grade
    'S': (2, [[.95, .2],
              [.05, .8]], ['I'], [2]),   # SAT
    'L': (2, [[.1, .4, .99],
              [.9, .6, .01]], ['G'], [3])   # Latter
}
prop_factor = 100  # Used to convert float probability to int number of outcomes

def make_bayes_network() -> BayesianNetwork:
    all_cpd = [
        TabularCPD(var, var_card, values, e_var, e_card)
        for var, (var_card, values, e_var, e_card) in net_config.items()]

    for cpd in all_cpd:
        print(f"[make_bayes_network] bayes_cpd:\n{cpd}")

    model = BayesianNetwork([('D', 'G'), ('I', 'G'), ('I', 'S'), ('G', 'L')])
    model.add_cpds(*all_cpd)

    return model

def make_relation_graph() -> RelationGraph:
    rgb = RelationGraphBuilder(
        variables={var: {f"{var}({i})" for i in range(0, card)} for var, (card, _, _, _) in
net_config.items()},
        relations={"r"})

    for var, (card, cpd, p_var, p_card) in net_config.items():
        for vs, prop in _value_map_from_cpd(var, card, p_var, p_card, cpd).items():
            outcome = _build_outcome(rgb, var, p_var, list(vs))
            count = int(prop * prop_factor)
            assert count == (prop * prop_factor), "[make_relation_graph] Select correct 'prop_factor'"
            print(f"[make_relation_graph] Outcome: {outcome}, count = {count}")
            rgb.add_outcome(outcome, count)

    rel_graph = rgb.build()
    print(f"[make_relation_graph] rel_graph:\n{rel_graph.print_samples()}")
    # rel_graph.visualize_outcomes()
    # rel_graph.folded_graph().visualize()

    return rel_graph

def comparing_joint_probability(bayes_net: BayesianNetwork, rel_graph: RelationGraph) -> None:
    joint_factor: DiscreteFactor = VariableElimination(bayes_net).query(variables=list(net_config.keys()))
    joint_graph: RelationGraph = rel_graph.make_joined()
    joint_factor_len = sum(1 for _ in product(*[range(card) for card in joint_factor.cardinality]))
    joint_graph_len = len(joint_graph.outcomes.items())

    print(f"[comparing_joint_probability] joint_factor ({joint_factor_len}):\n{joint_factor}")
    print(f"[comparing_joint_probability] Joint relation
```

```python
            graph({joint_graph_len}):\n{joint_graph.print_samples()}")
        # joint_rgraph.visualize_outcomes()

        assert joint_factor_len == joint_graph_len, \
            "Number of joint outcomes should be same as number rows in joint factor"

        factor_values = _value_map_from_factor(joint_factor)
        graph_values = _normalize(_value_map_from_graph(joint_graph))

        _compare_values_map(factor_values, graph_values)

        factor_marginals = _marginals_from_factor(joint_factor)
        graph_marginals = joint_graph.marginal_variables_probability()

        _compare_margin_map(factor_marginals, graph_marginals)

def comparing_inference(bayes_net: BayesianNetwork, rel_graph: RelationGraph) -> None:
    margin_factor_i, margin_graph_i = _run_inference_on(['I'], bayes_net, rel_graph)

    print(f"[comparing_inference] margin_factor_i = {margin_factor_i}")
    print(f"[comparing_inference] margin_graph_i = {margin_graph_i}")

    margin_graph_i.pop('I')
    _compare_margin_map(margin_factor_i, margin_graph_i)

    margin_factor_ig, margin_graph_ig = _run_inference_on(['I', 'G'], bayes_net, rel_graph)

    print(f"[comparing_inference] margin_factor_ig = {margin_factor_ig}")
    print(f"[comparing_inference] margin_graph_ig = {margin_graph_ig}")

    margin_graph_ig.pop('I')
    margin_graph_ig.pop('G')
    _compare_margin_map(margin_factor_ig, margin_graph_ig)

if __name__ == '__main__':
    bn = make_bayes_network()
    rg = make_relation_graph()
    comparing_joint_probability(bn, rg)
    comparing_inference(bn, rg)
```

## D.2. Comparing with Markov network (comparing_with_markov_network.py)

```python
nodes = ['A', 'B', 'C', 'D']
edges = [
    (['A', 'B'], [30,  5, 1,  10]),
    (['B', 'C'], [100, 1, 1, 100]),
    (['C', 'D'], [1, 100, 100, 1]),
    (['D', 'A'], [100, 1, 1, 100]),
]

joined_values = [
    (0, 0, 0, 0),
    (0, 0, 0, 1),
    (0, 0, 1, 0),
    (0, 0, 1, 1),
    (0, 1, 0, 0),
    (0, 1, 0, 1),
    (0, 1, 1, 0),
    (0, 1, 1, 1),
    (1, 0, 0, 0),
    (1, 0, 0, 1),
    (1, 0, 1, 0),
    (1, 0, 1, 1),
```

```python
        (1, 1, 0, 0),
        (1, 1, 0, 1),
        (1, 1, 1, 0),
        (1, 1, 1, 1)]

    def make_markov_network() -> MarkovNetwork:
        g = MarkovNetwork()
        g.add_nodes_from(nodes=nodes)
        g.add_edges_from(ebunch=[e[0] for e in edges])
        for edge, values in edges:
            f = DiscreteFactor(edge, cardinality=[2, 2], values=values)
            g.add_factors(f)
            print(f"Factor: \n {f}")
        return g

    def make_relation_graph() -> RelationGraph:
        rgb = RelationGraphBuilder(
            variables={n: {f"{n}(0)", f"{n}(1)"} for n in nodes},
            relations={"r"})
        for edge, values in edges:
            for sid, tid, i in [(0, 0, 0), (0, 1, 1), (1, 0, 2), (1, 1, 3)]:
                sn = f"{edge[0]}({sid})"
                tn = f"{edge[1]}({tid})"
                outcome = rgb.sample_builder()\
                    .add_relation({(edge[0], sn), (edge[1], tn)}, "r")\
                    .build()
                print(f"Outcome: {outcome}, count = {values[i]}")
                rgb.add_outcome(outcome, values[i])
        rel_graph = rgb.build()
        # rel_graph.visualize_outcomes()
        return rel_graph

    def comparing_variable_joint_probability(markov_net: MarkovNetwork, rel_graph: RelationGraph) -> None:
        ab_factor: DiscreteFactor = markov_net.factors[0].copy()
        bc_factor: DiscreteFactor = markov_net.factors[1].copy()
        cd_factor: DiscreteFactor = markov_net.factors[2].copy()
        da_factor: DiscreteFactor = markov_net.factors[3].copy()

        joint_markov: DiscreteFactor = ab_factor * bc_factor * cd_factor * da_factor
        joint_relation_graph: RelationGraph = rel_graph.make_joined()

        print(f"Joint markov:\n{joint_markov}")
        print(f"Joint relation graph:\n{joint_relation_graph.print_samples()}")
        # joint_relation_graph.visualize_outcomes()

        joint_markov.normalize()

        joint_markov_prop = _get_markov_prop(joint_markov)
        joint_relation_graph_prop = _get_relation_graph_prop(joint_relation_graph)

        print(f"Joint markov prop:         {_format_prop(joint_markov_prop)}")
        print(f"Joint relation graph prop: {_format_prop(joint_relation_graph_prop)}")

        assert joint_markov_prop == joint_relation_graph_prop, "Expect evaluated probabilities to be same"

    def comparing_inference(markov_net: MarkovNetwork, rel_graph: RelationGraph) -> None:
        ab_factor: DiscreteFactor = markov_net.factors[0].copy()
        bc_factor: DiscreteFactor = markov_net.factors[1].copy()
        cd_factor: DiscreteFactor = markov_net.factors[2].copy()
        da_factor: DiscreteFactor = markov_net.factors[3].copy()

        print(f"ab_factor =\n{ab_factor}")
        print(f"da_factor =\n{da_factor}")

        ab_factor.set_value(0, A=1, B=0)
```

```python
        ab_factor.set_value(0, A=1, B=1)
        da_factor.set_value(0, A=1, D=0)
        da_factor.set_value(0, A=1, D=1)

        print(f"After inference on E=A_0, ab_factor =\n{ab_factor}")
        print(f"After inference on E=A_0, da_factor =\n{da_factor}")

        joint_markov: DiscreteFactor = ab_factor * bc_factor * cd_factor * da_factor

        print(f"joint_markov =\n{joint_markov}")

        evidence = rel_graph.sample_builder().build_single_node("A", "A(0)")
        inference_on_a_0 = rel_graph.conditional_graph(evidence)

        print(f"Inference graph on A_0:\n{inference_on_a_0.print_samples()}")
        # inference_on_a_0.visualize_outcomes()

        joint_inference_graph: RelationGraph = inference_on_a_0.relation_graph().make_joined()

        print(f"Join inference graph on A_0:\n{joint_inference_graph.print_samples()}")
        # joint_inference_graph.visualize_outcomes()

        joint_markov_prop = _get_markov_prop(joint_markov)
        joint_inference_prop = _get_relation_graph_prop(joint_inference_graph.relation_graph())

        print(f"Inference E=A_0 markov prop = {_format_prop(joint_markov_prop)}")
        print(f"Inference E=A_0 relation graph prop = {_format_prop(joint_inference_prop)}")

        assert joint_markov_prop == joint_inference_prop, "Expect evaluated probabilities to be same"

if __name__ == '__main__':
    mn = make_markov_network()
    rg = make_relation_graph()
    comparing_variable_joint_probability(mn, rg)
    comparing_inference(mn, rg)
```